\documentclass[11pt,a4paper]{article}
\usepackage{times,latexsym}
\usepackage{url}
\usepackage[T1]{fontenc}
\usepackage{multirow}
\usepackage{tabu}
\usepackage{graphicx}
\usepackage{booktabs}
\usepackage{tablefootnote}
\usepackage[acceptedWithA]{tacl2021v1}
\usepackage{enumitem}
\usepackage[all]{nowidow}

\usepackage{xspace,mfirstuc,tabulary}

\newif\iftaclinstructions
\taclinstructionsfalse % AUTHORS: do NOT set this to true
\iftaclinstructions
\renewcommand{\confidential}{}
\renewcommand{\anonsubtext}{(No author info supplied here, for consistency with
TACL-submission anonymization requirements)}
\newcommand{\instr}
\fi

\iftaclpubformat % this "if" is set by the choice of options

\else

\fi
\title{Learning to Paraphrase Sentences to Different Complexity Levels}

\author{
  Alison Chi
  ,{ }
  Li-Kuang Chen
  ,{ } 
  Yi-Chen Chang$^*$
  ,{ } 
  Shu-Hui Lee\Thanks{Equal contribution}
  ,{ }
  Jason S. Chang
  \\
  \ \\
  National Tsing Hua University, Taiwan
  \\
  \texttt{achi@gapp.nthu.edu.tw},{ }
  \texttt{lkchen@gapp.nthu.edu.tw},{ }
  \\
  \texttt{yichen@nlplab.cc},{ }
  \texttt{shlee@nlplab.cc},{ }
  \texttt{jason@nlplab.cc}
  \\
}

\date{}

\begin{document}
\maketitle
\begin{abstract}
 While sentence simplification is an active research topic in NLP, its adjacent tasks of sentence complexification and same-level paraphrasing are not. 
To train models on all three tasks, we present two new unsupervised datasets. We compare these datasets, one labeled by a weak classifier and the other by a rule-based approach, with a single supervised dataset. Using these three datasets for training, we perform extensive experiments on both multitasking and prompting strategies. Compared to other systems trained on unsupervised parallel data, models trained on our weak classifier labeled dataset achieve state-of-the-art performance on the ASSET simplification benchmark. Our models also outperform previous work on sentence level targeting. Finally, we establish how a handful of Large Language Models perform on these tasks under a zero-shot setting.
\end{abstract}

\section{Introduction}  \label{sec:intro}
Paraphrasing a sentence to a targeted level of complexity is a natural language processing task that has not received much attention. Most work focuses solely on sentence simplification: decreasing the syntactic and lexical complexity of a sentence in order to make it easier to understand while preserving its original meaning \cite{siddharthan2002architecture, siddharthan2006syntactic, zhu-etal-2010-monolingual, woodsend2011learning, tacl-newsela, zhangLapata2017, alva2020}. This task has applications for second language (L2) learners and people with neural conditions that impede their reading comprehension abilities \cite{alva2020}. There has been limited work on sentence complexification, which is the exact opposite of sentence simplification: increasing the syntactic and lexical complexity of a given sentence \cite{berov-standvoss-2018-discourse}. 

As far as we know, there has not been any work done on same-level paraphrasing, which we define as paraphrasing a given sentence without changing its complexity level. However, all three tasks have important potential applications in computer-assisted language learning. 

Services like Grammarly\footnote{\url{https://www.grammarly.com}} and LinggleWrite \cite{tsai-etal-2020-lingglewrite} aim to correct grammatical and lexical writing errors, especially for L2 learners. Others aim to generate example usage sentences for new words \cite{example-suggester-2017}, as well as suggest potential paraphrases of learners’ sentences in order to improve the diversity of their writing \cite{nlplab-paraphrase-2015}. In addition to suggesting general paraphrase rewrites, the online writing assistant WordTune\footnote{\url{https://www.wordtune.com}} allows users to control both the length (correlated to complexity) and formality level of its paraphrase suggestions \cite{zhao2022leveraging}. 

Despite the existence of these paraphrasing systems commercially, to the best of our knowledge, there has been no academic work on paraphrasing to different complexity levels. Writing assistants and general language learning systems could benefit from this. A learner might want to see more concise ways of expressing their ideas (simplifications), more advanced or idiomatic ways of expressing them (complexifications), or suggestions that match their writing level (same-level paraphrases). We present models for all three tasks. For these tasks, we construct two automatically labeled (unsupervised) datasets and compare them to one human-labeled (supervised) dataset.  

Our first automatic labeling method is rule-based according to Flesch-Kincaid Grade Level (FKGL). FKGL can be calculated automatically as a weighted score consisting of sentence length and syllable information \cite{kincaid1975derivation}. A lower score means simpler output, and the lowest possible score is -3.40. Although this metric has been widely used for automatic evaluation of sentence simplification systems, it has been criticized for being easy to manipulate without increasing the simplification quality of the output \cite{tanprasert-kauchak-2021-flesch}.

Our second automatic labeling method is weak classification according to the six Common European Framework of Reference for Languages (CEFR) levels. The CEFR is used in standardized testing around the world to describe the language ability of L2 learners.\footnote{\url{https://www.cambridgeenglish.org/exams-and-tests/cefr}} It contains six levels in the order of increasing complexity: A1, A2, B1, B2, C1, and C2.\footnote{Levels that fall within the same letter are closer together than those that belong to different letters. E.g. A1 and A2 are more similar to each other than A1 and B1.} Unlike FKGL, the CEFR is based on a holistic combination of lexical, syntactic, and conceptual features and requires professionals to determine \cite{council2001common}. We construct a new, weakly labeled CEFR-annotated sentence and phrase dataset from the English Profile and Cambridge Dictionary, which we call CEFR-CEP (CEFR-Cambridge-English-Profile). We train a classifier to classify sentences and phrases into any of the six levels. 

From the ParaNMT dataset \cite{wieting-gimpel-2018-paranmt}, we create both CEFR-labeled and FKGL-labeled unsupervised sentence simplification, complexification, and same-level paraphrasing datasets. We also use a supervised dataset called Newsela-Auto \cite{jiang-etal-2020-neural}. On all three datsets, we fine-tune T5 models. We conduct ablation studies on multitasking configurations, comparing performance of single-task, two-task, and three-task models. We also compare two prompting strategies: absolute prompting, where we prepend target complexity level to the input sentence, and relative prompting, where we prepend level direction to the input sentence. Finally, we assess how Large Language Models (LLMs) perform on these tasks in a zero-shot setting. Our contributions are as follows:

\begin{itemize}[noitemsep,nolistsep]
    \item To our knowledge, we are the first to attempt the task of changing complexity level in any direction. From our in-depth fine-tuning experiments as well as a brief study on how well LLMs can change complexity, we establish new benchmarks.
    \item Our CEFR-labeled ParaNMT dataset produces state-of-the-art results on the ASSET simplification benchmark for models trained on unsupervised parallel data. 
    \item Our absolute prompting models outperform previous level targeting work on the Newsela-Manual benchmark.
    \item We release our ParaNMT data, CEFR classifier, and best fine-tuned paraphrasing models to the public.\footnote{\url{https://github.com/alisonhc/change-complexity}} We also release the CEFR-CEP test data used for human evaluation. The source dataset is publicly available on EVP, EGP, and Cambridge websites and can be obtained via their data request process.\footnote{\url{https://languageresearch.cambridge.org/academic-research-request-form}}
\end{itemize}

\section{Related Work}
\subsection{Sentence Complexity Classification} \label{subsec:CEFR-work}
Much work has been done on complexity level classification as a component of Automatic Readability Assessment, but it has mostly focused on the document level \cite{xia-etal-2016-text, lee-etal-2021-pushing} and not the sentence level due to a shortage of sentence-level datasets. In English, data from Newsela,\footnote{\url{https://newsela.com}} which contains articles that have been manually simplified to four different target levels, has been widely used \cite{tacl-newsela, lee-etal-2021-pushing, lee-vajjala-2022-neural}. Newsela sentences levels can be automatically derived for sentence-level research. However, since Newsela levels (US grade ranges) are per document, not every sentence level corresponds to its document's level. The OneStopEnglish corpus \cite{2018-onestopenglish}, which consists of sentences and documents labeled at three ESL levels, is also widely used. Since readability is highly subjective and dependent on a specific audience or set of standards, it is difficult to apply a single readability assessment scheme to a variety of domains. \citet{lee-vajjala-2022-neural}'s pairwise ranking model has made progress on this, demonstrating strong accuracy on out-of-domain (OOD) data.

As the CEFR is a widely used international standard, readability classification into CEFR levels has been attempted \cite{xia-etal-2016-text, khallaf2021, arase-etal-2022-cefr}. But most of this work has focused on documents, collections of documents, and individual words \cite{kerz2021, schmalz2021, settles2020, gaillat2022}. There is a very limited amount of work on sentence level classification \cite{volodina2013, khallaf2021, arase-etal-2022-cefr}. \citet{arase-etal-2022-cefr} present CEFR-SP, the first human-labeled CEFR English sentence-level dataset, sourcing sentences from Newsela-Auto and Wiki-Auto \cite{jiang-etal-2020-neural} in addition to the Sentence Corpus of Remedial English (SCoRE).\footnote{\url{https://www.score-corpus.org/en}}  A BERT classifier trained on CEFR-SP achieves 84.5\% F1 on the in-domain test set \cite{arase-etal-2022-cefr}. 

\subsection{Changing Sentence Complexity} 
\label{subsec:complexity-work}
Most work in changing sentence complexity focuses on lowering sentence level to specific grades. The Newsela corpus \cite{tacl-newsela, jiang-etal-2020-neural} has been used to train controlled simplification models to target level \cite{scarton2018, agrawal2019, nishihara2019, kew-ebling-2022-target, tani-etal-2022-benchmark}. To our knowledge, there have been three previous attempts at sentence complexification, also known as text or discourse embellishment. \citet{berov-standvoss-2018-discourse} introduce the task and train a LSTM on a story corpus and the inverse of a simplification corpus, WikiLarge, which contains aligned sentence pairs from English and Simple English Wikipedia articles \cite{zhangLapata2017}. \citet{naskar-etal-2019-text} also use WikiLarge. And more recently, \citet{sun2023teaching} train BART \cite{lewis-etal-2020-bart} on reversed simplification sentence pairs from Newsela. There has been no previous work on same-level paraphrasing.

\subsection{Sentence Simplification} \label{subsec:simplification-work}
{\bf Supervised Data}
Many sentence simplification systems adopt the architecture of machine translation, requiring complex-simple sentence pairs to train \cite{zhu-etal-2010-monolingual, wubben2012sentence, narayan2014hybrid, zhangLapata2017, alva2020}. WikiLarge \cite{zhangLapata2017}, described in Section \ref{subsec:complexity-work}, has been widely used. Models trained on this dataset can be easily applied to test sets that source their data from Wikipedia such as ASSET \cite{alva2020asset} and the Turk Corpus \cite{xu2016sari}. Newsela, also described in Section \ref{subsec:complexity-work}, has been a popular source for sentence simplification datasets \cite{tacl-newsela, zhangLapata2017}. \citet{jiang-etal-2020-neural} present a sentence alignment model to generate the larger datasets of Wiki-Auto and Newsela-Auto. Their human annotators also developed the smaller Newsela-Manual dataset. Although most of the aforementioned corpora contain sentences that are automatically aligned, they are still considered supervised because the text was simplified by humans.

{\bf Unsupervised Data}
Since there are few supervised datasets, methods have been proposed to generate unsupervised datasets, which often consist of mined paraphrases. Backtranslation, or translating a sentence into a language and then back into the original language, has been used to generate paraphrases \cite{lu-etal-2021-unsupervised-method}. Other work has used heuristics like embedding similarity to mine semantically similar sentence pairs \cite{martin2020muss}. An effective way of training on unsupervised parallel data is the use of control tokens to allow models to hone in on features that correlate with sentence simplicity. For example, the ACCESS method prepends tokens that specify output length, similarity of output and input, output word rank, and output tree depth to the beginning of each input sentence \cite{martin2020access}. As these tokens are by default prepended in plain text \textbf{before tokenization}, they are functionally a form of prompt learning. 

{\bf Multitask Learning}
Multitask learning has proven useful for overcoming lack of data and improving simplification quality. Entailment \cite{guo2018}, paraphrase generation \cite{guo2018, maddela-etal-2021-controllable}, copy prediction \cite{maddela-etal-2021-controllable}, translation \cite{agrawal2019, mallinson2020}, and summarization \cite{dmitrieva2020} have all been used as auxiliary tasks for simplification models. It has been shown in the past that training a model on multiple very similar tasks can improve its performance on each individual task \cite{ratner2018snorkel, liu2019multi}. Although simplification, complexification, and same-level paraphrasing belong to the same general task of changing sentence complexity, training a multitask model with all three has not previously been attempted. The use of prompts for both training and inference has proven particularly useful for multitasking with pretrained models. \citet{scialom-etal-2022-fine} fine-tune a T5 model with eight new tasks, including sentence simplification, with prompts either prepended to the input text or embedded as part of a template depending on the task.

{\bf Inference with Large Language Models}
Research has been done on whether LLMs can simplify text without further training. \citet{feng2023sentence} show that GPT-3.5-Turbo\footnote{\url{https://openai.com/blog/chatgpt}} produces a SARI score of 44.67 for zero-shot prompting and 47.06 for single-shot prompting, surpassing previous state-of-the-art scores. 
\citet{Ryan_2023} find that BLOOM \citep{workshop2023bloom} achieves high meaning preservation and fluency but fails to simplify as well as smaller fine-tuned models.
\citet{aumiller-gertz-2022-unihd} use an ensemble of prompts on GPT-3 \citep{brown-etal-2020-gpt3} producing state-of-the-art results for lexical simplification specifically.

\section{CEFR Level Classification} \label{sec:CEFR-classifier}
In order to automatically label paraphrase data with complexity levels, we first train a sentence level classification model. In theory, any of the few English sentence-level readability datasets can be used for training. However, CEFR-SP \cite{arase-etal-2022-cefr} and Newsela \cite{tacl-newsela} may contain data that we use for training and testing our later paraphrasing models, so we do not use either of those. The other option of OneStopEnglish \cite{2018-onestopenglish} has very few sentence pairs, and upon inspection, we find its simplest level to appear more complex than CEFR A1. Therefore, we create a new CEFR-labeled corpus for our needs, CEFR-CEP.

\subsection{Data} \label{subsec:CEFR-classifier-data}
We combine data from the English Profile and Cambridge Dictionary.\footnote{\url{https://dictionary.cambridge.org/zht}} Our main source, English Profile \cite{capel2012}, contains CEFR levels that map to word senses or grammar concepts. It contains two searchable databases, English Vocabulary Profile (EVP)\footnote{\url{https://www.englishprofile.org/wordlists/evp}} and English Grammar Profile (EGP).\footnote{\url{https://www.englishprofile.org/english-grammar-profile/egp-online}} 
\begin{table}[t]
%   \begin{center}
    \begin{center}
    \resizebox{0.8\columnwidth}{!}{%
    \begin{tabular}{@{}l|ll@{}}
    \toprule
      \multirow{3}{*}{\textbf{Source Distribution}}
        & EVP: 32079\\
      \multirow{3}{*}{}
        & EGP: 3620\\
      \multirow{3}{*}{}
        & Cambridge Dict: 3714\\
      \midrule
      \multirow{6}{*}{\textbf{Level Distribution}}
        & A1: 1790\\
      \multirow{6}{*}{}
        & A2: 3890\\
      \multirow{6}{*}{}
        & B1: 7445\\
      \multirow{6}{*}{}
        & B2: 10558\\
      \multirow{6}{*}{}
        & C1: 5921\\
      \multirow{6}{*}{}
        & C2: 9809\\
      \midrule
      \multirow{2}{*}{\textbf{Sentence vs. Phrase}}
        & Sentence Count: 28638\\
      \multirow{2}{*}{}
        & Phrase Count: 10775\\
         \bottomrule
    \end{tabular}%
    }
    \end{center}
        \caption{CEFR-CEP information} \label{table:level-classifier-dataset-statistics}
\end{table}

\begin{figure}[t]
    \centerline{\includegraphics[width=\columnwidth]{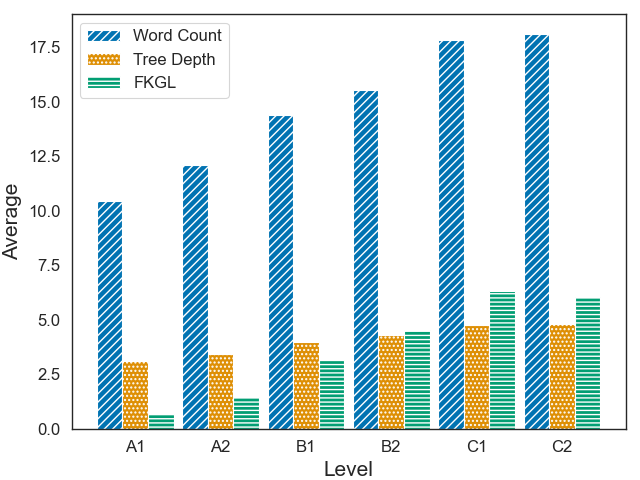}}
    \caption{For texts in CEFR-CEP, the average word count, tree depth, and FKGL per CEFR level.}
    \label{fig:level-distribution}
\end{figure}

Each entry in EVP corresponds to a word, and each of its possible definitions (word senses) is marked with its CEFR level along with one or more example usage sentences or phrases from either a real learner or a dictionary. EVP words, but not example sentences, have been used in the past to create lexical simplification datasets \cite{uchida2018cefr, fujinuma2021cefr}. EGP and the Cambridge Dictionary are structured similarly to EVP, containing CEFR levels and examples for grammar concepts and word senses respectively. We automatically label these EVP, EGP, and dictionary examples with their entries' CEFR levels. We eliminate any duplicates from our combined dataset. Further details about CEFR-CEP are shown in Table \ref{table:level-classifier-dataset-statistics}. 

This method assumes that for each word sense or grammar concept, its example sentences/phrases match its CEFR level. This is likely false some of the time. However, analysis on the CEFR-CEP sentences shows that our assumed CEFR levels correlate strongly with other metrics associated with sentence complexity: word count, tree depth, and FKGL, as shown in Figure \ref{fig:level-distribution}.

\subsection{Model}   \label{subsec:classifier-training}
On CEFR-CEP, we train a BERT classifier \cite{devlin-etal-2019-bert} in addition to SVM and LSTM baselines, with an 80-10-10 train-validation-test split.
The BERT-base-cased [CLS] token embedding serves as the sentence representation and the input to our classifier, which is made up of one linear layer and trained with cross-entropy loss like in previous work \cite{arase-etal-2022-cefr}. 
Its outputs are softmax probabilities for each of the six CEFR levels, and we use an Adam optimizer \cite{adam-kingma} with the best learning rate of 3e-5.\footnote{On a single NVIDIA GPU, we use the AllenNLP library \cite{gardner-2018-allennlp} to train for three epochs with a batch size of 32.}

In addition to the BERT model, we train two baselines on the same data. The first is a Support Vector Machine (SVM) classifier with Term Frequency-Inverse Document Frequency (TF-IDF) for its embeddings and a Radial Basis Function kernel \cite{scholkopf1997comparing}. We use the optimal cost and gamma hyperparameters of 10 and 1 respectively. We also train a LSTM classifier with a single dense layer and Word2Vec Google News vectors \cite{mikolov-word2vec-gnews} as its embedding layer, Adam optimization with an optimal learning rate of 4e-3, softmax activation, and cross entropy loss.

\subsection{Evaluation}    \label{subsec:classifier-evaluation}
We perform automatic evaluation on our held-out CEFR-CEP test data with four evaluation metrics. Our F1 scores are weighted to take label imbalance into account. 

\begin{itemize}[noitemsep,nolistsep]
    \item \textbf{6-Level F1 (6-F1)}: The prediction F1 for the six CEFR levels.
    \item \textbf{3-Level F1 (3-F1)}: The prediction F1 for the three CEFR levels A, B, and C.
    \item \textbf{Adjacent Accuracy (Adj-Acc)}: the percentage where the prediction’s deviation from the test label is less than or equal to one.\footnote{Under this metric, a prediction of A2 would be considered accurate if the test label was A1, A2, or B1, because the deviation from A2 is one or less.}
    \item \textbf{Mean Absolute Error (MAE)}: a number between 0 and 5. The average amount that the prediction deviates from the test label.\footnote{Prediction 0 (A1), test label 1 (A2) corresponds to MAE of 1. Prediction 1, test label 5 (C2) corresponds to MAE of 4.}
\end{itemize}

Table \ref{table:cefr-classifier-results} shows the results for each metric on the baseline and BERT models. For every metric, the BERT model performs better. But 6-F1 is only 59.78\%, and we posit that it is so difficult to get an exact match with dataset CEFR level because of dataset flaws mentioned in Section \ref{subsec:CEFR-classifier-data}: namely that we label each example text according to the level of its corresponding word sense or grammar concept, which is not always correct. But Adj-Acc is a high value of 90.64\%, showing that our model has very close estimation, and the low MAE of 0.52 is consistent with this. Our SVM baseline scores similarly to the LSTM despite having much more information-rich embeddings. 

\begin{table}[t]
\begin{center}
\resizebox{0.8\columnwidth}{!}{%
\begin{tabular}{@{}l|ccccc@{}}
\toprule
\textbf{Model} & \textbf{6-F1$\uparrow$} & \textbf{3-F1$\uparrow$} & \textbf{Adj-Acc$\uparrow$} & \textbf{MAE$\downarrow$} \\
\midrule \midrule
SVM & 57.40 & 71.29 & 80.54 & 0.68 \\
\midrule
LSTM & 53.17 & 70.00 & 82.04 & 0.71 \\
\midrule
\midrule
BERT & \textbf{59.78} & \textbf{76.80} & \textbf{90.64} & \textbf{0.52} \\
\bottomrule
\end{tabular}%
}
\end{center}
\caption{CEFR classifier results on CEFR-CEP test set} \label{table:cefr-classifier-results}
\end{table}

Since we will use our classifier to add CEFR labels to the OOD ParaNMT dataset, we conduct a study to see to what extent its labels match human labels on the ParaNMT data. On our preprocessed ParaNMT set (see Section \ref{subsec:paraphrase-data} for details), we sample 60 sentence pairs: 20 where their classified levels are the same and 40 where their classified levels differ by at least two (e.g. A2-B2 but not A2-B1). We split the different-level pairs into two groups, simplification where the higher level sentence comes first and complexification where the lower level one does. We then ask four native English speakers to examine each sentence pair and label which sentence is simpler: the first, the second, or neither. \textbf{These three labels map to the categories of complexification, simplification, and same-level paraphrasing respectively.} 

Inter-rater agreement, or nominal Krippendorff's Alpha \cite{krippendorff2011computing}, is a fairly low $0.27$, where $0$ means no agreement (chance) and $1$ means perfect agreement. Because we want to evaluate on only reliable labels, we just consider the sentence pairs where three or more of the raters agree. These amount to 39 out of 60 pairs with agreement of $0.48$. We test both our CEFR classifier and FKGL on these 39 gold labels.

\begin{table}[t]
\begin{center}
\resizebox{0.7\columnwidth}{!}{%
\begin{tabular}{l|ll}
\toprule
\textbf{Category} & \textbf{CEFR F1} & \textbf{FKGL F1} \\
\toprule \midrule
  Simplification & \textbf{53.33} & 46.15   \\ 
Complexification &  50.0 & \textbf{64.52}    \\ 
Same Level & 12.50  & \textbf{28.57}   \\
\bottomrule
\end{tabular}%
}
\end{center}
\caption{F1 of CEFR classifier vs. FKGL predictions on 39 human labels} \label{table:human-evaluation-cefr}
\end{table}

We compare our CEFR classifier's predictions with those of FKGL. Table \ref{table:human-evaluation-cefr} shows the F1 of the CEFR versus FKGL methods on the gold labels for each of the three categories of simplification, complexification, and same-level paraphrasing. FKGL performs better for classifying complexification and same-level paraphrasing, while CEFR classification performs better for simplification. However, F1 is universally low, casting doubt on the reliability of our weak labeling approaches. Our gold human labels are also potentially problematic: only six of the 60 sentence pairs that were rated as same-level paraphrasing met our criterion of three out of four raters agreeing, compared to 15 and 18 for simplification and complexification respectively. From these results, we tentatively hypothesize that sentence simplification models trained on data labeled by the CEFR classifier will perform better than those trained on FKGL-labeled data, while complexification and same-level paraphrasing models trained FKGL-labeled data will perform better than those trained on CEFR-labeled data.

\section{Paraphrasing Data} \label{sec:paraphrase-data}
Next, we construct datasets for simplification, complexification, and same-level paraphrasing. Details are included in Table \ref{table:paraphrase-data-details}.

\begin{table}[t]
  \begin{center}
% \centering
    \resizebox{0.9\columnwidth}{!}{%
    \begin{tabular}{l|ll}
      \toprule % <-- Toprule here
      \textbf{Dataset} & \textbf{Tasks} & \textbf{Size} \\
      \midrule \midrule % <-- Midrule here
      Newsela-Auto & Simplification & 238,597  \\
      & Complexification & 238,662 \\
      \midrule
      ParaNMT-CEFR & Simplification & 1,287,794  \\
      & Complexification & 1,287,795  \\ 
    & Same Level & 1,287,795 \\
      \midrule
      ParaNMT-FKGL & Simplification & 1,287,794 \\
      & Complexification & 1,287,794 \\
      & Same Level & 1,287,794 \\
      \bottomrule % <-- Bottomrule here
    \end{tabular}%
    }
  \end{center}
      \caption{Paraphrasing dataset details}
    \label{table:paraphrase-data-details}
\end{table}

\subsection{Supervised Data}
Our supervised data source is Newsela-Auto,\footnote{Request data at \url{https://newsela.com/data}.} a sentence simplification corpus derived from Newsela news articles targeted at five levels and written by education professionals, where level $0$ is the complex original and 1-4 are simplifications of increasing degree \cite{tacl-newsela}. Their sentences must be aligned to create a sentence pair corpus from these original articles. Previous methods have aligned using metrics like Jaccard similarity \cite{zhangLapata2017}. Newsela-Auto's pairs are aligned according to a neural CRF model \cite{jiang-etal-2020-neural}, and its pairs are more numerous (666k) and creatively rewritten than previous Newsela alignments.\footnote{To stay consistent with previous work, we employ the same train-test-validation split.} Newsela-Auto does not contain level labels, so we use string matching with the original Newsela to find each sentence's level \cite{tacl-newsela}.\footnote{Due to the limitations of this retroactive approach, our resulting corpus is slightly smaller than the original: 394,108 instead of 394,300 for training and 43,305 instead of 43,317 for validation.}

A limitation of Newsela-Auto and other simplification datasets like WikiLarge \cite{zhangLapata2017} and Wiki-Auto \cite{jiang-etal-2020-neural} is that they are only meant to contain different-level pairs. Therefore, we only conduct simplification and complexification experiments on this dataset. For the two-task dataset, we flip the order of exactly half of the sentence pairs. For the two single-task datasets, we extract all simplification and complexification pairs from the two-task dataset but perform an additional filtering step of removing all pairs that were labeled as the same level according to our retroactive labeling algorithm. These pairs only number into a few thousand and are not enough to train a comparable same-level paraphrasing model.

\subsection{Unsupervised Data} \label{subsec:paraphrase-data}
To contrast with our supervised dataset and fill the gap of missing same-level paraphrase pairs, we create two unsupervised datasets. We use ParaNMT, one of the largest paraphrase pair datasets available to the public, with 50 million sentence pairs generated through backtranslation of the Czeng1.6 corpus \cite{wieting-gimpel-2018-paranmt}. It contains data sourced from movie and educational video subtitles, European legislation proceedings, and medical websites \cite{bojar2016czeng}. ParaNMT has been used for sentence simplification in the past \cite{martin2020muss}.

To determine our filtering techniques, we inspect samples from the corpus and find pairs that are identical or almost identical, very different in meaning, or that contain incomplete sentences. To alleviate these problems, we remove pairs where one sentence is contained in the other or where any sentence has less than three words. 

To encourage our models not to directly copy the input sentence, a problem that occurs in both sentence simplification \cite{dong-etal-2019-editnts} and paraphrase generation \cite{thompson-post-2020}, we only include aggressive paraphrases. We remove pairs where Sentence-BERT cosine similarity \cite{reimers2019sbert} is below 60\% or above 80\%. From our observations, these thresholds exclude pairs that are different in meaning or too similarly phrased. 

We want ParaNMT-CEFR and and ParaNMT-FKGL to be as similar as possible for the sake of comparison. From our filtered data, we use the CEFR classifier to label the level of each sentence. To maximize the likelihood that a level difference between the two sentences exists (see Table \ref{table:cefr-classifier-results}'s Adj-Acc), we only select pairs where the level difference is two or greater.\footnote{E.g. we keep A1-B1 pairs but remove A2-B1 pairs.} For the same-level dataset, we select pairs where the sentences are classified as exactly the same level.\footnote{E.g. A1-A1 but not A1-A2}

We are left with 2,575,589 different-level pairs and 6,207,876 same-level pairs. For both the CEFR-based and FKGL-based labeling schemes, \textbf{we derive all of our simplification, complexification, and same-level paraphrasing data from these two sets}. For ParaNMT-CEFR, we halve the different-level dataset and re-order it to create one simplification and one complexification dataset. We then sample from the same-level pairs to get an equal-sized same-level set. To create ParaNMT-FKGL, we calculate the FKGL of each sentence (rounded to two decimal points). If the FKGL of the two sentences in a pair differs at all, we consider it a different-level pair. If it is \textbf{exactly the same}, we consider it a same-level pair. We are able to derive 65.16\% of our different-level pairs from the ParaNMT-CEFR different-level set. The other 878,449 are taken from the ParaNMT-CEFR same-level pairs. We sample from the resulting data to match ParaNMT-CEFR's in size. The train-validation-test split is 80-10-10 for both ParaNMT datasets. We have made these data available to the public.\footnote{\url{https://github.com/alisonhc/change-complexity}} 

\section{Paraphrasing Experiments} \label{sec:paraphrasing-experiments}
We train models on the three tasks of sentence simplification, sentence complexification, and same-level paraphrasing. We train ablations for training dataset (Newsela-Auto, ParaNMT-CEFR, ParaNMT-FKGL), multitasking configuration (1-3 tasks), and prompting strategy (relative/absolute). Including our baselines, we train 42 models in total. See Table \ref{table:task-prompts} for details. 

\begin{table}[t]
  \begin{center}
% \centering
    \small
    % \resizebox{\columnwidth}{!}{%
    \begin{tabular}{l|l}
      \toprule % <-- Toprule here
      \textbf{Task} & \textbf{Prompt(s)} \\
      \midrule \midrule % <-- Midrule here
      Simplification & "level down: "  \\
      & "change to level X: "  \\
      \midrule
      Complexification & "level up: "  \\
      & "change to level X: "  \\ 
      \midrule
      Same-level & "same level: " \\
      \bottomrule % <-- Bottomrule here
    \end{tabular}%
    % }
  \end{center}
      \caption{Prompt(s) for each task. For same-level paraphrasing single-task models, we only train REL prompt ablations. For simplification, complexification, and all two-task and three-task configurations, both REL and ABS prompt ablations are trained.}
    \label{table:task-prompts}
\end{table}

\subsection{Models}
For all models, we use a single NVIDIA GPU, a batch size of 32 after gradient accumulation, and maximum decoding length of 300 tokens. We fine-tune 34 ablations on \textbf{T5} \cite{raffelT5}, a pre-trained transformer\footnote{We fine-tune T5-base with the transformers library \cite{wolf-2020-transformers}. After 3 epochs, we automatically select the model checkpoint with the lowest validation loss.}
We also perform limited experiments with Flan-T5-base \cite{chung2022scaling}, a more recent instruction-tuned version of T5. We train ParaNMT-CEFR single-task and 2-task simplification and complexification ablations (6 models). However, since we find in Section \ref{subsubsec:flan-t5-results} that it does not perform as well as T5, we focus our main experiments on T5.

\subsection{Prompting Strategies}
\label{subsec:prompting-strategies}
At inference time, we prepend the corresponding prompt to the beginning of each input sentence, as this strategy was used for T5 \cite{raffelT5}.

{\bf Relative}
Simplification, complexification, and same-level paraphrasing correspond exactly to the prompts \emph{"level down: "}, \emph{"level up: "}, and \emph{"same level: "}. We train on the data of one, two, or all three tasks, adding the corresponding task prompt to the front of each input sentence. We call this relative (REL) prompting because the prompt denotes the relative difference between the levels of the input and output sentence: down, up, or same. This scheme has 7 possible task combinations.

{\bf Absolute}
For each task combination besides single-task same-level paraphrasing, we use prompts that specify absolute (ABS) output level. For training, we insert \emph{"change to level X: "}, where $X$ is the level of the output.\footnote{For ParaNMT-CEFR, $X$ is the CEFR level A/B/C. For ParaNMT-FKGL, $X$ is FKGL rounded to two decimal points. And for Newsela-Auto, $X$ is one of the Newsela levels 0-4.}
ABS prompting theoretically has an advantage over REL prompting because we can change the prompt to match the level of a test dataset's output sentence. To compare the two prompting strategies on equal footing, we remove this advantage. With the exception of Section \ref{subsec:level-targeting}, for ABS prompting inference, \textbf{we use the same prompt for every test input no matter the output level}. Therefore, we can only evaluate these models on simplification and complexification and not on same-level paraphrasing. 

\subsection{Baselines}
We train paraphrasing baselines, the first trained on the entire ParaNMT-CEFR dataset and the other trained on ParaNMT-FKGL. Each dataset consists of one third simplification data, one third complexification data, and one third same-level paraphrasing data, but at train time, we use the prompt \emph{"paraphrase: "} for each input.\footnote{We also train an LSTM baseline per task per ParaNMT dataset using REL prompting, but we do not report the results because they do not add to the analysis.}

\section{Paraphrasing Evaluation}    \label{sec:paraphrase-evaluation}
We perform both automatic and human evaluation. To compare all 40 experiment models, we only report automatic evaluation results. We perform human evaluation on just one model per task.

\subsection{Automatic Evaluation}   \label{subsec:automatic-evaluation}
We first discuss each individual task. Then, we discuss our ablation results more generally.

{\bf Evaluation metrics}
We report SARI and FKGL.\footnote{We use the EASSE Python library to compare with previous sentence simplification research \cite{alva-manchego-easse}.}
\begin{itemize}[noitemsep,nolistsep]
    \item \textbf{SARI} (System output Against References and against the Input sentence) is the most important automatic metric for text simplification. Ranging from zero to 100, it represents the F1 for a model’s added, kept, and deleted n-grams when comparing the input and reference sentences \cite{xu2016sari}. 
    \item \textbf{FKGL} (Flesch–Kincaid Grade Level) is a weighted score with sentence length and syllable information \cite{kincaid1975derivation}. It was introduced in Section \ref{sec:intro}. We consider the best FKGL score to be that closest to the gold reference FKGL in a given test set.
\end{itemize}
{\bf Evaluation data}
For simplification and complexification, we use the ASSET and Newsela-Manual test sets. These simplification benchmarks can be easily reversed for the complexification task. There are no existing benchmarks that can be straightforwardly applied to same-level paraphrasing. Therefore, we use sentence pairs from the ParaNMT corpus. In all tables and figures, we denote task type to u/d/s for up (complexification), down (simplification), and same. 

\begin{itemize}[noitemsep,nolistsep]
    \item \textbf{ASSET} has 359 test sentences, each with 10 human-written reference sentences \cite{alva2020asset}. For simplification, we use this dataset as-is. For complexification, we consider each reference sentence to be an input and the corresponding test sentence to be an output, resulting in 3590 one-to-one pairings.
    \item \textbf{Newsela-Manual} contains Newsela sentence pairs where each pair is annotated as \emph{aligned}, \emph{partially aligned}, or \emph{not aligned} \cite{jiang-etal-2020-neural}.\footnote{There is no overlap between Newsela-Auto training or validation data and Newsela-Manual test data.} We collect all \emph{aligned} and \emph{partially aligned} pairs and follow \citet{kew-ebling-2022-target}'s method to automatically fix the alignments between \emph{partially aligned} pairs. We include pairs from all input levels to all output levels and remove pairs where the output is an exact copy of the input, resulting in 2,748 pairs.\footnote{For simplification, we use the dataset as-is, and for complexification, we reverse it.} 
    \item \textbf{Newsela-Manual by Level} contains sentences where the complex level 0 maps to each of the simpler levels 1-4. To evaluate our models' level targeting ability, we use the same configuration as \citet{kew-ebling-2022-target}, which does not filter out input-output copies. We also create a complexification version where the simple input is level 4 and the four possible outputs are levels 3-0.
\item \textbf{ParaNMT-s} Since there is no publicly available same-level paraphrasing dataset, we sample from both the FKGL and CEFR versions of the ParaNMT-same set to collect 128,779 pairs.\footnote{There is no overlap between our resulting test set and either of the training or validation sets.} This corpus is inherently noisy due to its unsupervised nature. We hope that in future work, a cleaner same-level paraphrasing dataset with human labels will be available. 
\end{itemize}

% ASSET + Newsela manual simplification
\begin{table}[t]
\centering
\resizebox{\columnwidth}{!}{%
\begin{tabular}{@{}l|cc|cc}

\toprule
\multicolumn{1}{l}{}      & 
\multicolumn{2}{c}{ASSET}       & 
\multicolumn{2}{c}{Newsela-Manual}   \\
\cmidrule(lr){2-5}
\textbf{Model} &
  \textbf{SARI$\uparrow$} &
  \textbf{FKGL} &
    \textbf{SARI$\uparrow$} &
  \textbf{FKGL} \\
\midrule \midrule
\textbf{Baselines}  & \multicolumn{4}{c}{}   
\\
\midrule
Reference  & 44.89  & 6.49   & -- & 5.80  \\
T5-CEFR-Para & 39.58 & 9.88 & 36.13 & 9.73 \\
T5-FKGL-Para & 39.45 & 9.90 & 36.0 & 9.69 \\
\midrule \midrule
\textbf{Unsupervised Data}  & \multicolumn{4}{c}{}   
\\
\midrule

MUSS-mined
& 42.65      & 8.23      & 38.80  & 7.26 \\
\citealt{lu-etal-2021-unsupervised-method}  & 42.69  & 7.94      & -- & --     \\
\midrule
T5-CEFR-u-d-ABS (B)  & \textbf{43.65}  &  7.91   & 39.13 & 8.09  \\  
T5-CEFR-d-s-ABS (B) & 43.45 & 8.51 & \textbf{39.67} & 8.24 \\  
T5-FKGL-d-ABS (see caption) & 42.38   & \textbf{7.03}  & 37.81 & 2.47  \\
T5-FKGL-d-s-ABS (3.0) & 42.31 & 6.81 & 39.29 & \textbf{5.90} \\  
\midrule \midrule
\textbf{Supervised Data}  & \multicolumn{4}{c}{} \\
\midrule
MUSS-wiki-mined & \textbf{44.15} & \textbf{6.05}   & 41.38 & 6.67       \\
\citealt{clive2021control} & 43.58   & 5.97  & -- & --  \\
\midrule
T5-News-d-ABS (4) & 40.87   & 5.96    & 41.54 & \textbf{5.76}  \\
T5-News-u-d-REL & 39.97   & 5.92    & \textbf{42.44} & 5.91  \\
\bottomrule
\end{tabular}%
}
\caption{\textbf{Simplification} on ASSET and Newsela-Manual. Models abbreviated to [Model]-[Data]-[Tasks]-[ABS or REL prompting]. MUSS-mined and MUSS-wiki-mined come from \citet{martin2020muss}. MUSS and \citet{clive2021control} use ACCESS prompting \cite{martin2020access}. \citet{lu-etal-2021-unsupervised-method} create their own corpus via backtranslation. For ABS models, we enclose in parentheses the target level we used for prompting at inference time. T5-FKGL-d-ABS uses 3.0 for ASSET and 0.0 for Newsela-Manual.}
\label{table:level-down-table-asset-newsela} 
\end{table}

% ASSET + Newsela manual complexification
\begin{table}[t]
\centering
\resizebox{\columnwidth}{!}{%
\begin{tabular}{@{}l|cc|cc}
\toprule
\multicolumn{1}{l}{}      & 
\multicolumn{2}{c}{ASSET}       & 
\multicolumn{2}{c}{Newsela-Manual}   \\
\cmidrule(lr){2-5}
\textbf{Model} &
  \textbf{SARI$\uparrow$} &
  \textbf{FKGL} & 
    \textbf{SARI$\uparrow$} &
  \textbf{FKGL} \\
\midrule \midrule
\textbf{Baselines} \\
\midrule
Reference  & --  & 10.46 & -- & 10.14       \\
T5-CEFR-Para  & 42.09  & 7.46   & 39.41 & 6.92       \\
T5-FKGL-Para  & 42.28  & 7.40   & 39.83 & 6.93       \\
\midrule \midrule
\textbf{Unsupervised Data}  & \multicolumn{4}{c}{} \\
\midrule
MUSS-mined      & 44.06 & 7.92  & 38.46 & 7.85       \\
\midrule
% CEFR paranmt
T5-CEFR-u-ABS (C) & 43.87   & 7.79    & \textbf{40.98} & 7.61  \\
T5-CEFR-u-s-ABS (C) & 43.44 & 7.70 & 39.60 & 7.50 \\ 
T5-FKGL-u-ABS (12.0) & 43.86 & 13.76 & 40.21 & 12.36 \\
T5-FKGL-u-s-ABS (11.0) & \textbf{44.07} & \textbf{11.87} & 40.32 & \textbf{11.10} \\
% newsela
\midrule \midrule
\textbf{Supervised Data}  & \multicolumn{4}{c}{} \\
\midrule
MUSS-wiki-mined   & \textbf{42.51} & 7.89  & 37.97 & 7.40       \\
\citet{sun2023teaching} & 40.0 & 8.30 & -- & -- \\
\midrule
T5-News-u-ABS (0) & 38.96 & \textbf{9.82}  & \textbf{42.21} & \textbf{9.46}       \\
T5-News-u-REL & 36.90 & 8.10  & 42.07 & 7.64       \\
\bottomrule
\end{tabular}
}
\caption{\textbf{Complexification} on ASSET and Newsela Manual. See Table \ref{table:level-down-table-asset-newsela}'s caption for naming details. We obtained model weights and data for \citet{sun2023teaching}'s ComplexBART model and ran inference ourselves. However, since their Newsela training data overlaps with the Newsela-Manual test set, we only report ASSET scores for ComplexBART.} \label{table:level-up-table-asset-newsela} 
\end{table}

\subsubsection{Simplification Results}
We report results in Table \ref{table:level-down-table-asset-newsela} on both the ASSET and Newsela-Manual test sets. Besides baselines, we divide the table into two sections, one for models trained on unsupervised data and the other for supervised data. We only report our two best performing ablations per training dataset. For ABS prompting CEFR and Newsela-Auto (News) models, we try all possible prompts. For FKGL models, we try a range of prompts (0.0-7.0) and pick the best ones. For MUSS models, which are open source \cite{martin2020muss}, we report their best scores on ASSET and do our own parameter search on the Newsela-Manual validation set to derive optimal prompts. On both benchmarks, all models outperform baselines in SARI score. We achieve a new state-of-the-art for unsupervised parallel data, with the highest SARI score of 43.65 on ASSET going to T5-CEFR-u-d-ABS (prompt B).\footnote{We say unsupervised \textbf{parallel} data because ChatGPT, mentioned in Section \ref{subsec:simplification-work}, has a higher score \cite{feng2023sentence}.} 
Our supervised model T5-News-u-d-REL has the highest SARI score on the Newsela-Manual benchmark, outperforming baselines and MUSS.

\subsubsection{Complexification Results}
We report SARI and FKGL on reversed ASSET and reversed Newsela-Manual. Table \ref{table:level-up-table-asset-newsela} contains results arranged in the same way as for simplification. For FKGL prompts, we try a range of 10.0-17.0. We do a grid search to find MUSS parameters. MUSS-mined almost matches our best performing model's SARI on ASSET, even beating its supervised data counterpart and \citet{sun2023teaching}'s ComplexBART. But it falls much shorter of our best models on Newsela-Manual.

Between ParaNMT-CEFR and ParaNMT-FKGL models, the latter produce the highest SARI on ASSET and highest FKGL on both test sets. 
However, after inspecting model outputs, we find that for every FKGL model whose SARI surpasses our highest ParaNMT-CEFR SARI score, the outputs contain many degenerate repetitions. For example, consider the ASSET input simple sentence \textbf{The state capital is Aracaju.} T5-CEFR-u-s-ABS with prompt C produces the slightly longer sentence \begin{quote}the capital of the state is Aracaju.\end{quote} But T5-FKGL-u-s-ABS with prompt 11.0 produces a 295 word output starting with \begin{quote}the capital of the state is Aracaju, the capital of the state is the capital of the state of the state of\end{quote} MUSS SARI also surpasses ParaNMT-CEFR on ASSET. However, their outputs contain fewer degenerate repetitions according to an inspection of the outputs. We believe this quality difference is due to problems with the ParaNMT dataset that are exacerbated by organizing it by FKGL score, a length-based metric. The MUSS-mined training data contains human-written sentences that were mined according to similarity metrics \cite{martin2020muss}. ParaNMT, on the other hand, is the result of machine translation \cite{wieting-gimpel-2018-paranmt}, which can sometimes enter repetitive loops during decoding \cite{holtzman2019curious, welleck2019neural}. Future work on backtranslation datasets could attempt to filter out sentences that contain these repetitions.

 We also find that degenerate repetitions are not adequately captured by SARI, which only counts \textbf{unique n-grams} that are added, kept, and deleted compared to the gold references \cite{xu2016sari, alva-manchego-easse}. This means that \textbf{as long as a model’s repetitions have added no or very few unique new words to the sentence, they will not be reflected in SARI}. Therefore, we suggest that for sentence complexification, a modified SARI should be used that takes word counts into consideration. We leave this to future work.

% ASSET + Newsela auto complexification
\begin{table}[t]
% \centering
\begin{center}
\resizebox{0.7\columnwidth}{!}{%
\begin{tabular}{@{}l|cc}
\toprule
\textbf{Model} &
  \textbf{SARI$\uparrow$} &
  \textbf{FKGL} \\
\midrule \midrule
\textbf{Baselines} \\
\midrule
Reference  & --  & 2.82       \\
T5-CEFR-Para  & \textbf{49.40}  & 2.76       \\
T5-FKGL-Para  & 48.21  & \textbf{2.82}       \\
\midrule 
\textbf{Experiment Models}  & \multicolumn{2}{c}{} \\
\midrule
% CEFR paranmt
T5-CEFR-u-d-s-REL  & 48.26   & 2.86      \\
T5-FKGL-u-d-s-REL  & 45.75   & 2.90      \\
\bottomrule
\end{tabular}%
}
\end{center}
\caption{\textbf{Same-level paraphrasing} on ParaNMT-s. See Table \ref{table:level-down-table-asset-newsela}'s caption for naming details.}
\label{table:same-level-table} 
\end{table}

\begin{figure*}[t]
\centering
    \centerline{\includegraphics[scale=0.45]{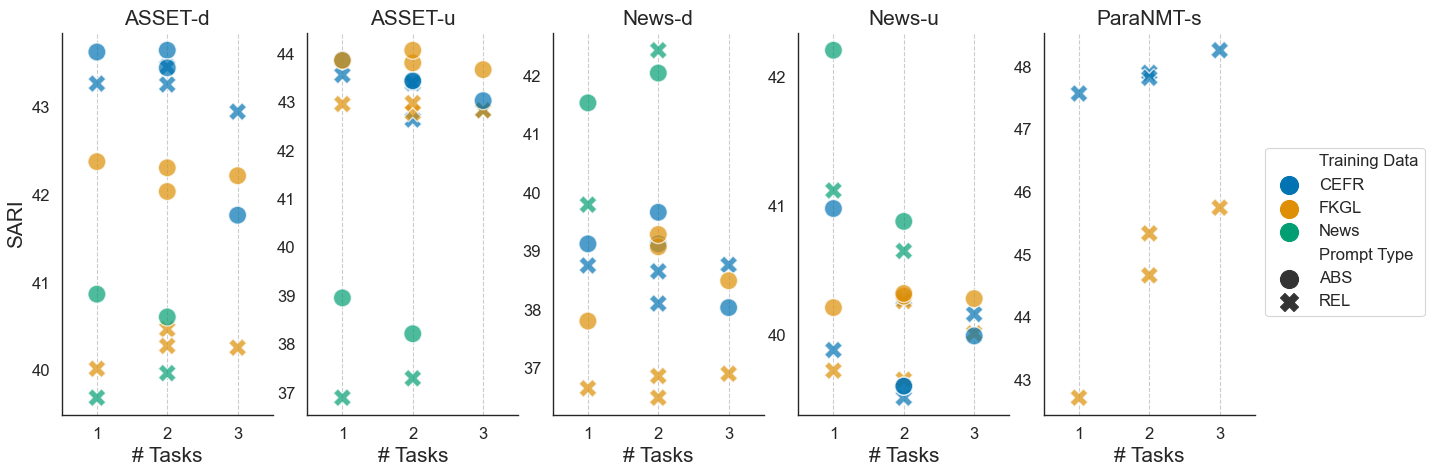}}
    \caption{All ablation results. Tasks abbreviated as u (up, complexification), d (down, simplification), and s (same, same-level paraphrasing). ASSET-d and News-d correspond to the original ASSET and Newsela-Manual sets. The -u indicates that they were reversed for complexification.}
    \label{fig:all-results-plot}

\end{figure*}

\subsubsection{Same-level Paraphrasing Results}
In Table \ref{table:same-level-table}, we report results for all of our baselines along with our best performing CEFR and FKGL models. Notably, both of our CEFR and FKGL paraphrasing \emph{baselines} outperform their corresponding experiment models, which were trained on the \textbf{exact same data}, the only difference being prompting strategy. When we compare T5-CEFR-Para’s outputs with those of T5-CEFR-u-d-s-REL, we find that after tokenization, the former copies the input 4.40\% of the time, while the latter does so 10.42\% of the time. The ParaNMT-s test set copies input 0.31\% of the time after tokenization. Since we are unable to perform a quantitative human evaluation comparing these outputs, we are left with two possible theories.

The first is that our T5 paraphrasing baselines are actually learning to same-level paraphrase. When presented with data where a third increases level, a third decreases level, and a third keeps level the same, the model picks the average option, which is same-level paraphrasing. The second theory is that the sentences in our same-level paraphrasing data are not actually the same level. After all, both our CEFR and FKGL methods in Section \ref{subsec:classifier-evaluation} have extremely low F1 on human labels for same-level paraphrasing: 12.5\% for CEFR and 28.57\% for FKGL. However, we doubt this theory because of our positive human evaluation results (see Section \ref{subsec:human-evaluation}). 

\begin{table}[t]
\centering
\small
\resizebox{\columnwidth}{!}{%
\begin{tabular}{@{}l|cc|cc}
\toprule
\multicolumn{1}{l}{ } & 
\multicolumn{2}{c}{Simplification (d)} & 
\multicolumn{2}{c}{Complexification (u)}  \\
\cmidrule(lr){2-5}
\textbf{Model} &
  \textbf{ASSET} &
  \textbf{News} & 
  \textbf{ASSET} &
  \textbf{News} \\
\midrule \midrule
Best T5-CEFR & \textbf{43.65} & \textbf{39.67} & \textbf{43.87} & \textbf{40.98} \\
\midrule
CEFR-ABS-d (B) & \textbf{42.91} & \textbf{39.28} & -- & -- \\
CEFR-ABS-u (see caption) & -- & -- & \textbf{42.84} & \textbf{40.57} \\
CEFR-ABS-u-d (d-B, u-C) & 42.45 & 38.81 & 42.33 & 39.46 \\
\midrule
CEFR-REL-d & 42.46 & 38.75 & -- & -- \\
CEFR-REL-u & -- & -- & 42.73 & 40.55 \\
CEFR-REL-u-d & 42.64 & 38.79 & 42.12 & 39.66 \\
\bottomrule
\end{tabular}
}
\caption{Flan-T5 SARI for all trained ablations. For ABS models, the best prompt(s) are shown in parenthesis. CEFR-ABS-u uses B for ASSET and C for News.}
\label{table:flan-t5-results} 
\end{table}
\subsubsection{Flan-T5 Results} \label{subsubsec:flan-t5-results}
Table \ref{table:flan-t5-results} shows SARI for the six Flan-T5 ablations we trained along with the best SARI scores from our T5 experiments on the same dataset of ParaNMT-CEFR. Interestingly, the best Flan-T5 scores never surpass the best from T5. And when directly comparing scores for each ablation, \textbf{T5 outperforms Flan-T5 for 12 out of the 16 cases}.

This may be surprising, as Flan-T5 performs better on a variety of tasks and benchmarks for zero- and few-shot inference \cite{chung2022scaling}. But Flan-T5 has \textbf{not} been shown to be better than T5 for fine-tuning on new datasets. We suspect that the reason for its degraded performance compared to T5 is that fine-tuning incurs catastrophic forgetting, diminishing the benefits gained from its previous instruction-tuning. While \citet{scialom-etal-2022-fine} report that T5 models can continually learn new tasks without catastrophic forgetting, rehearsal \citep{rehearsal-shin-etal-2017} is still required for the models to retain their previously learned skills.

\subsubsection{Ablation Study Results} \label{subsubsec:ablation-results}
Figure \ref{fig:all-results-plot} shows results for all T5 experiment models on all test sets, the x-axis being number of tasks per model and the y-axis being SARI score. Each data point is annotated with task combination.  

{\bf Multitasking}
There is no clear winner among multitasking configurations. Single and two-task models often perform better than three-task ones, with the exception of same-level models, where SARI increases with the number of tasks. 
Many high-scoring two-task models were trained on tasks that are not opposite (i.e. u-s and d-s but not u-d). However, for simplification, the highest scoring models for ASSET and Newsela-Manual were both trained on the u-d ablation. For T5-News-u-d-REL, this is not noteworthy because REL prompts are distinct for each task (see Table \ref{table:task-prompts}). But strikingly, T5-CEFR-u-d-ABS scores best on ASSET with prompt \textbf{B} even though in theory, upon seeing the middle prompt B (as opposed to A or C), the model should not know whether to increase or decrease a sentence's complexity. Upon further investigation, we find that the reason for this is likely that the training dataset contains approximately double the amount of $C \rightarrow B$ simplifications as $A \rightarrow B$ complexifications.

{\bf Prompt type}
For FKGL models, ABS prompting always performs better than REL prompting. For News models, ABS prompting performs better in all but one case. For CEFR models, results are mixed, but ABS prompting performs slightly better on average.
Compared to CEFR and Newsela levels, FKGL is very fine-grained, with up to two decimal point precision. The fact that FKGL models always perform better for ABS prompting than for REL, while CEFR and News models do not, suggests that using prompts that contain very fine-grained output information might improve performance. 
Additionally, among just single-task models, ABS prompting always performs best, but this strategy is favored less and less as the amount of tasks increases. This indicates that using a more complex prompting strategy incurs a greater performance cost as the number of tasks increases.

{\bf Data labeling scheme}
As expected, models trained on Newsela-Auto perform better on Newsela-Manual than models trained on ParaNMT data. However, they mostly fail to achieve as high of SARI on non-Newsela data as ParaNMT models achieve on Newsela data, and they are some of the worst performing models on ASSET. For ABS prompting, FKGL models often outperform CEFR models on complexification, but for REL prompting, FKGL models almost universally do worse. For same-level paraphrasing, it is notable that ParaNMT-CEFR models have much higher SARI than ParaNMT-FKGL ones despite the fact that the ParaNMT-s test dataset is half ParaNMT-CEFR and half ParaNMT-FKGL. This, and the fact that complexification FKGL model outputs contain degenerate repetitions that SARI does not reflect, shows that the CEFR method is the most robust automatic labeling method. Future work could experiment with finer-grained CEFR labels (6, not 3) and less fine-grained FKGL labels (intervals instead of two decimal precision). 

\begin{table}[t]
% \centering
\begin{center}
\resizebox{\columnwidth}{!}{%
\begin{tabular}{@{}l|cc||l|cc}
\toprule
% \cmidrule(lr){2-3}
\multicolumn{3}{c}{\textbf{Simplification}} 
& \multicolumn{3}{c}{\textbf{Complexification}}  \\
\midrule
\textbf{Target Level} & $\textbf{SARI$\uparrow$}$ & \textbf{FKGL} 
& \textbf{Target Level} & $\textbf{SARI$\uparrow$}$ & \textbf{FKGL}\\

\midrule \midrule

$0$ $\rightarrow$  $ 1$  & -- &   $9.05$  
& 4 $\rightarrow$ 3  & -- & 5.46 \\ \midrule
MUSS & 38.71 & 7.34 
& MUSS & 35.14 & 6.24 \\
Ours & \textbf{39.81} & \textbf{10.30} 
& Ours & \textbf{41.82}  & \textbf{4.90}\\ \midrule

0 $\rightarrow$ 2  & -- &  7.13  
& 4$\rightarrow$ 2 & -- & 7.05\\ \midrule
MUSS & \textbf{42.37} & \textbf{7.06}
& MUSS & 37.25 & \textbf{6.22} \\
Ours & 41.81 & 7.82 
& Ours & \textbf{40.97}   & 5.90 \\ \midrule

0 $\rightarrow$ 3  & -- & 5.51  
& 4 $\rightarrow$ 1 & -- & 9.06 \\ \midrule
MUSS & 40.21 & \textbf{4.88} 
& MUSS & 37.19 & 6.10 \\
Ours & \textbf{44.81} & 6.31  
& Ours & \textbf{41.52}  & \textbf{6.85}\\ \midrule

0 $\rightarrow$ 4  & --  &  3.89 
& 4 $\rightarrow$ 0  & -- & 11.46 \\ \midrule
MUSS & 40.08 & \textbf{4.64} 
& MUSS & 34.53 & 5.80 \\
Ours & \textbf{46.77} & 4.83
& Ours & \textbf{42.44}  & \textbf{8.33}\\
\bottomrule
\end{tabular}%
}
\end{center}
\caption{Level targeting for \textbf{simplification} and \textbf{complexification} on Newsela-Manual. We compare our scores to supervised MUSS \cite{martin2020muss}. Our simplification model is T5-News-u-d-ABS. For each level, we display reference FKGL. See Table \ref{table:level-down-table-asset-newsela} for naming conventions.} \label{table:level-target-table} 
\end{table}

\subsubsection{Level Targeting Results} \label{subsec:level-targeting}
Table \ref{table:level-target-table}  show our Newsela-Auto models’ abilities to target specific levels for simplification and complexification. For brevity, we show results from only one of our models per table along with the best previous work baseline, supervised MUSS \cite{martin2020muss}, for which we derive optimal parameters via grid search. 
For every level, our models achieve higher SARI than previous work, with the exception of $0$ $\rightarrow$ $2$ simplification, where MUSS wins. However, it appears that our models are better at targeting aggressive simplifications and complexifications than slight ones: SARI generally increases as target level deviates further from input level. 
 The results from Section \ref{subsubsec:ablation-results} show that even when we are not using ABS prompting to its full strength, it often surpasses REL prompting in performance. These level-targeting results confirm that ABS prompting at its full strength does better. 

\subsection{Human Evaluation}   \label{subsec:human-evaluation}
We carry out a human evaluation on all three tasks. We use a 1-5 Likert scale across three separate categories: task performance, meaning preservation, and fluency. Due to limited resources, we choose just one model per task. 
We choose models ParaNMT models for our evaluation.
For simplification, T5-CEFR-u-d-ABS with prompt B scores best on ASSET, but due to the prompt B task ambiguity discussed in Section \ref{subsubsec:ablation-results}, we choose T5-CEFR-d-ABS with prompt B, which scores second best with a SARI of \textbf{43.63}. For complexification, we use the highest scoring CEFR model, T5-CEFR-u-ABS with prompt C, even though some of the FKGL models have higher SARI scores on ASSET. This is because, as mentioned in Section \ref{subsec:automatic-evaluation}, FKGL models produce numerous degenerate repetitions that do not hurt SARI score. Finally, for same-level paraphrasing, we choose T5-CEFR-u-d-s-REL because of its highest SARI score on ParaNMT-s.

Due to limited human evaluation resources, out of the three tasks, we only compare our simplification model to a baseline. We choose supervised MUSS \cite{martin2020muss}, a publicly available state-of-the-art model that we also used in Section \ref{subsec:automatic-evaluation}. We use its best performing ASSET prompts. So as to directly compare the three tasks of simplification, complexification, and same-level paraphrasing on the exact same dataset, something not done in Section \ref{subsec:automatic-evaluation}, we do not use a benchmark simplification dataset. We instead source data from the CEFR-CEP test set, which our paraphrasing models have not seen and our CEFR classifier has not been trained or validated on. However, because of this choice, there are no reference paraphrases to compare model outputs to, preventing us from using a reference baseline. We do not use any baseline because in the absence of a single one that fits all three tasks, it would require dramatically more labeling work. 

\begin{table}[t]
\resizebox{\columnwidth}{!}{%
\begin{tabular}{@{}l|ccc@{}}
\toprule
 &
  \textbf{Task} &
  \textbf{Meaning} &
  \textbf{Fluency}  \\ \toprule \midrule
  \textbf{Simplification} \\ \midrule
MUSS & $2.96_{\pm 0.23} $         & $3.63_{\pm 0.34}$          & $4.71_{\pm 0.15}$   \\ 
\hspace{0.7cm}\small{Agreement} &
  \small{0.33} &
  \small{0.63} &
  \textbf{\small{0.28}}  \\
  
Ours   & $\textbf{3.04}_{\pm 0.26}$         & $\textbf{4.24}_{\pm 0.27}$          & $\textbf{4.74}_{\pm 0.14}$     \\ 
\hspace{0.7cm}\small{Agreement} &
  \small{\textbf{0.44}} &
  \small{0.60} &
  \small{0.26} \\
  
  \midrule
\textbf{Complexification} & $2.35_{\pm 0.23}$          & $4.12_{\pm 0.33}$          & $4.64_{\pm 0.14}$   \\ 
\hspace{0.7cm}\small{Agreement} &
  \small{0.28} &
  \textbf{\small{0.77}} &
  \small{0.18} \\
  
\midrule
\textbf{Same Level} & $\textbf{3.85}_{\pm 0.18}$ & $\textbf{4.72}_{\pm 0.15}$ & $\textbf{4.77}_{\pm 0.11}$ \\ 
\hspace{0.7cm}\small{Agreement} &
  \small{0.01} &
  \small{0.52} &
  \small{0.16} \\
\bottomrule
\end{tabular}%
}
\caption{Human evaluation results. Each row contains a mean rating from 1 to 5 with a confidence interval, plus inter-rater agreement below it.}
\label{table:human-evaluation-paraphrase}
\end{table}

From CEFR-CEP, we sample 13 sentences from each level A2-C1, amounting to 52 sentences that we release to the public.\footnote{\url{https://github.com/alisonhc/change-complexity}} We exclude A1 and C2 because simplifying or complexifying those sentences may not have an effect. We then run each of the four models on these sentences, producing 208 outputs. Three native English speakers each rate all outputs.\footnote{The raters are not told which outputs are from our models and which are not.} For each output, we average the ratings of the three evaluators. We then take the 95\% confidence interval across each model’s rating category along with inter-rater agreement using ordinal Krippendorff's Alpha \cite{krippendorff2011computing}, a number between zero (random agreement) and one (perfect agreement). 

Table \ref{table:human-evaluation-paraphrase} shows our results. For simplification, our model performs better than MUSS across all categories, especially meaning preservation. Across tasks, fluency is universally very high. This is a testament to the quality of these fine-tuned language models. Agreement is highest for meaning preservation, perhaps the most objective metric. We find that task performance is lowest for complexification, which is consistent with our intuition that this is the most difficult task, demanding the most additions and leaving the most room for error. Finally, same-level paraphrasing has the highest scores out of 5 compared to the other tasks, likely because it requires the least amount of modification. This is particularly interesting because of the fact that our paraphrasing baseline T5-CEFR-Para outperformed this model according to SARI on ParaNMT-s, calling into question whether the task models were effective at all. We told our raters to dock task performance points when a model exactly copied its input, but upon inspection of their ratings, we find that this is very inconsistent. So, this may be why inter-rater agreement is extremely low for task performance.

\begin{table}[t]
  \begin{center}
% \centering
    \small
    \resizebox{\columnwidth}{!}{%
    \begin{tabular}{l|l|l}
      \toprule % <-- Toprule here
      \textbf{Task} & \textbf{REL Prompt} & \textbf{ABS Prompt} \\
      \midrule \midrule % <-- Midrule here
      Simplification & "Please rewrite the & "Please rewrite the  \\
      or & following text to a & following text so \\
      Complexification & [less/more] advanced & that its [CEFR/FKGL] \\
      &  English level: " & level is X: "  \\
      \midrule
      Same-level & "Please rewrite the & -- \\
      & following text & \\
      & to the same & \\
      & English level: " & \\
      \bottomrule % <-- Bottomrule here
    \end{tabular}%
    }
  \end{center}
      \caption{Prompt(s) for each task. For CEFR ABS prompting, we use A for simplification and C for complexification. For FKGL ABS prompting, in two point intervals, we try levels 0-6 for simplification and 8-14 for complexification.}     \label{table:llm-prompts}
\end{table}

\section{Can LLMs Change Complexity Level?} \label{sec:llm_inference}
In this section, we perform an exploratory investigation into the simplification, complexification, and same-level paraphrasing abilities of LLMs.
\subsection{Experiments} \label{subsec:llm_experiments}
\subsubsection{Data} For simplification and complexification, we use ASSET like in Section \ref{subsec:automatic-evaluation}.\footnote{We do not use Newsela-Manual because we were not able to obtain clarity on whether sending data through OpenAI's API violates Newsela's licensing agreement.} For same-level paraphrasing, we randomly sample 400 sentence pairs from ParaNMT-s.\footnote{Cutting down on the original 128,779 pairs reduces both API costs and inference time.} 
% ASSET + Newsela manual complexification
\begin{table*}[t]
\centering
% \resizebox{\linewidth}{!}{%
\scalebox{0.72}{
\begin{tabular}{@{}l|cc|cc|cc}
\toprule
\multicolumn{1}{l}{}      & 
\multicolumn{2}{c}{Simplification (d)}       & 
\multicolumn{2}{c}{Complexification (u)}  & 
\multicolumn{2}{c}{Same Level (s)}   \\
\cmidrule(lr){2-7}
\textbf{Model} &
  \textbf{SARI$\uparrow$} &
  \textbf{FKGL} & 
    \textbf{SARI$\uparrow$} &
  \textbf{FKGL} &
      \textbf{SARI$\uparrow$} &
  \textbf{FKGL} \\
\midrule \midrule
Best Fine-tuned T5 & 43.65 & \textbf{7.03} & \textbf{44.07} & 9.82 & \textbf{48.26} & \textbf{2.86} \\
\midrule
GPT-3.5-Turbo (d-A, u-8) & \textbf{45.76} & 8.28 & \textbf{42.84} & \textbf{10.72} & \textbf{41.73} & 4.98 \\
GPT-NeoX-20B  (d-2, u-REL) & 35.85 & 5.77 & 34.78 & 3.89 & 34.52 & 2.43 \\
Flan-UL2 (d-4, u-10) & 32.50 & 4.91 & 34.58 & 5.51 & 21.85 & \textbf{2.73}   \\
Flan-T5-xxl (d-A, u-10) & 28.99 & 1.47 & 30.25 & 6.79 & 20.79 & 2.63  \\
OPT-IML-MAX-1.3B (d-0, u-8)  & 36.26 & \textbf{6.01} & 33.52 & 3.98 & 31.07 & 0.0   \\
\bottomrule
\end{tabular}
% }
}
\caption{LLM results based on best SARI per model, tested on ASSET. For u and d, best prompts are included in the Model column. Reference FKGL is 6.49 for simplification, 10.46 for complexification, and 2.82 for same-level paraphrasing.} \label{table:llm-results} 
\end{table*}
\subsubsection{Models}
For all models, we set temperature to 1.0 and limit output length to 50 tokens. We run inference in a zero-shot setting and leave an investigation into more sophisticated inference settings to future work. Due to hardware limitations, we are unable to run inference for models with more than 20 billion parameters. We mostly select instruction-tuned models because we expect them to do better with new tasks and prompts. We select five: GPT-3.5-Turbo,\footnote{\url{https://platform.openai.com/docs/model-index-for-researchers}} GPT-NeoX-20B \cite{black-etal-2022-gpt}, Flan-UL2 \cite{tay2023ul}, Flan-T5-xxl \cite{chung2022scaling}, and OPT-IML-MAX-1.3B \cite{iyer2023optiml}.

\subsubsection{Prompts}
Like in our fine-tuning experiments, we attempt both ABS and REL prompting. However, in this case, we construct prompts with more descriptive wording to better fit the zero-shot setting. Table \ref{table:llm-prompts} shows the prompts for each task. To determine them, we try different wording with GPT-3.5-Turbo to check for obvious differences in behavior. We find that for complexification, explicitly telling the model to "increase the complexity" of a piece of text produces undesirably long outputs, but the wording "advanced English level" does not. We keep terminology consistent across prompts.

\subsection{Results and Discussion} \label{subsec:llm_results} 

\begin{figure}[t]
\centering
    \centerline{\includegraphics[scale=0.45]{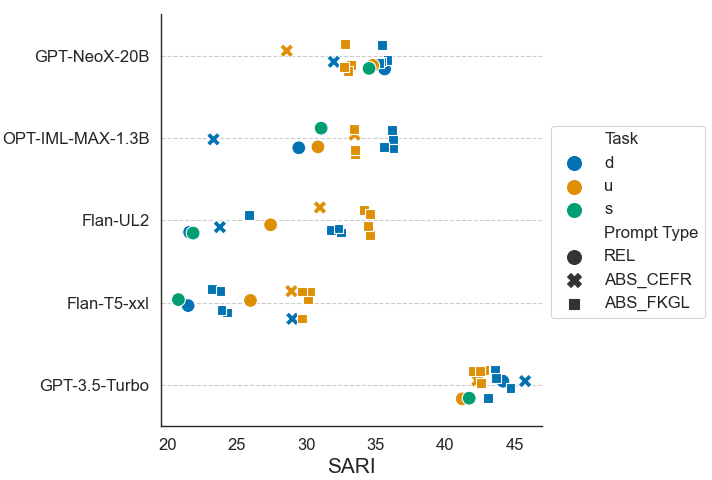}}
    \caption{All SARI scores per model, task, and prompt.}
    \label{fig:llm-results-plot}
\end{figure} 

Table \ref{table:llm-results} shows results for each LLM and task, and Figure \ref{fig:llm-results-plot} shows SARI for each LLM per task and prompt type. On all tasks, GPT-3.5-Turbo outperforms the rest of the models by a large margin. 
None of the other models produce SARI scores that come close to the paraphrasing baselines from Tables \ref{table:level-down-table-asset-newsela}, \ref{table:level-up-table-asset-newsela}, and \ref{table:same-level-table}, much less the fine-tuned T5 scores.  
We confirm this by inspecting model outputs: all besides GPT-3.5-Turbo contain hallucinations. For example, in response to CEFR prompting (and FKGL to a lesser degree), Flan-T5-xxl and Flan-UL2 often return a single letter instead of a sentence as the output, while OPT-IML-MAX-1.3B and GPT-NeoX-20B attach discussions of the CEFR to their outputs.

Despite the fact that the ABS prompting outputs contain more hallucinations than those from REL prompting, Figure \ref{fig:llm-results-plot} shows that ABS prompting generally produces higher SARI, echoing our findings from the fine-tuning experiments. For GPT-3.5-Turbo in particular, the ABS-CEFR prompt produces outputs with \textbf{higher} SARI for simplification than \citet{feng2023sentence}'s  REL prompting score of \textbf{44.67} in the zero-shot setting.

Notably, although GPT-3.5-Turbo outperforms our fine-tuned models on simplification, it does not on complexification, demonstrating the difficulty of the task. Models perform the worst at same-level paraphrasing, but this may be due to the unsupervised same-level dataset being worse in quality than supervised ASSET. 

The huge gap in performance between GPT-3.5-Turbo and the other models may be in part due to its size of 176B parameters being much larger than the next largest size of 20B. However, there is no obvious pattern regarding model size for the other four: for example, the smallest model of OPT-IML-MAX-1.3B performs competitively with the two 20B-parameter models. 

\section{Conclusion}    \label{sec:conclusion}
In this paper, we provide a general investigation of the task of changing sentence complexity, with thorough fine-tuning experiments and brief experiments with LLMs. For sentence simplification, our models surpass or are comparable to state-of-the-art systems. For sentence complexification and same-level paraphrasing, we set new benchmarks. We show that weak classification is an effective way to create strong unsupervised datasets and that target level absolute prompting is more effective than level direction relative prompting.   

This research leaves opportunities for future work. For example, using a stronger level classifier to label paraphrase data might improve performance for the paraphrasing tasks. In the same vein, different filtering of ParaNMT or another paraphrasing dataset \cite{hu2019parabank} could potentially be used. 
A human-labeled same-level paraphrasing test dataset does not yet exist, and a modified SARI metric that adequately penalizes repetitions is needed for sentence complexification. Our methods focus on English data, but they can be easily applied to other languages if a different classifier is trained \cite{khallaf2021, vasquez-rodriguez-etal-2022-benchmark} and a non-English paraphrasing dataset is used \cite{martin2020muss, scherrer2020tapaco, lu-etal-2021-unsupervised-method}. Finally, a thorough investigation on how well LLMs can change sentence complexity is necessary.

\iftaclpubformat

\section*{Acknowledgments}
We thank the reviewers and editor Dr. Sara Rosenthal for providing valuable feedback that made this paper much better. We would also like to thank Dr. Laura Vásquez-Rodríguez and Jhih-Jie Chen for their helpful advice, as well as Andrew Cavicchi for lending us compute power. Finally, we thank those who provided assistance with our human evaluation.
\else
\fi

\bibliography{tacl_change_level}

\begin{thebibliography}{82}
\expandafter\ifx\csname natexlab\endcsname\relax\def\natexlab#1{#1}\fi

\bibitem[{Agrawal and Carpuat(2019)}]{agrawal2019}
Sweta Agrawal and Marine Carpuat. 2019.
\newblock \href {https://doi.org/10.18653/v1/D19-1166} {Controlling text
  complexity in neural machine translation}.
\newblock In \emph{Proceedings of the 2019 Conference on Empirical Methods in
  Natural Language Processing and the 9th International Joint Conference on
  Natural Language Processing (EMNLP-IJCNLP)}, pages 1549--1564, Hong Kong,
  China. Association for Computational Linguistics.

\bibitem[{Alva-Manchego et~al.(2020{\natexlab{a}})Alva-Manchego, Martin,
  Bordes, Scarton, Sagot, and Specia}]{alva2020asset}
Fernando Alva-Manchego, Louis Martin, Antoine Bordes, Carolina Scarton,
  Beno{\^\i}t Sagot, and Lucia Specia. 2020{\natexlab{a}}.
\newblock \href {https://doi.org/10.18653/v1/2020.acl-main.424} {{ASSET}: {A}
  dataset for tuning and evaluation of sentence simplification models with
  multiple rewriting transformations}.
\newblock In \emph{Proceedings of the 58th Annual Meeting of the Association
  for Computational Linguistics}, pages 4668--4679, Online. Association for
  Computational Linguistics.

\bibitem[{Alva-Manchego et~al.(2019)Alva-Manchego, Martin, Scarton, and
  Specia}]{alva-manchego-easse}
Fernando Alva-Manchego, Louis Martin, Carolina Scarton, and Lucia Specia. 2019.
\newblock \href {https://doi.org/10.18653/v1/D19-3009} {{EASSE}: Easier
  automatic sentence simplification evaluation}.
\newblock In \emph{Proceedings of the 2019 Conference on Empirical Methods in
  Natural Language Processing and the 9th International Joint Conference on
  Natural Language Processing (EMNLP-IJCNLP): System Demonstrations}, pages
  49--54, Hong Kong, China. Association for Computational Linguistics.

\bibitem[{Alva-Manchego et~al.(2020{\natexlab{b}})Alva-Manchego, Scarton, and
  Specia}]{alva2020}
Fernando Alva-Manchego, Carolina Scarton, and Lucia Specia. 2020{\natexlab{b}}.
\newblock \href {https://doi.org/10.1162/coli_a_00370} {Data-driven sentence
  simplification: Survey and benchmark}.
\newblock \emph{Computational Linguistics}, 46(1):135--187.

\bibitem[{Arase et~al.(2022)Arase, Uchida, and Kajiwara}]{arase-etal-2022-cefr}
Yuki Arase, Satoru Uchida, and Tomoyuki Kajiwara. 2022.
\newblock \href {https://aclanthology.org/2022.emnlp-main.416} {{CEFR}-based
  sentence difficulty annotation and assessment}.
\newblock In \emph{Proceedings of the 2022 Conference on Empirical Methods in
  Natural Language Processing}, pages 6206--6219, Abu Dhabi, United Arab
  Emirates. Association for Computational Linguistics.

\bibitem[{Aumiller and Gertz(2022)}]{aumiller-gertz-2022-unihd}
Dennis Aumiller and Michael Gertz. 2022.
\newblock \href {https://aclanthology.org/2022.tsar-1.28} {{U}ni{HD} at
  {TSAR}-2022 shared task: Is compute all we need for lexical simplification?}
\newblock In \emph{Proceedings of the Workshop on Text Simplification,
  Accessibility, and Readability (TSAR-2022)}, pages 251--258, Abu Dhabi,
  United Arab Emirates (Virtual). Association for Computational Linguistics.

\bibitem[{Berov and Standvoss(2018)}]{berov-standvoss-2018-discourse}
Leonid Berov and Kai Standvoss. 2018.
\newblock \href {https://doi.org/10.18653/v1/W18-6603} {Discourse embellishment
  using a deep encoder-decoder network}.
\newblock In \emph{Proceedings of the 3rd Workshop on Computational Creativity
  in Natural Language Generation ({CC}-{NLG} 2018)}, pages 11--16, Tilburg, the
  Netherlands. Association for Computational Linguistics.

\bibitem[{Black et~al.(2022)Black, Biderman, Hallahan, Anthony, Gao, Golding,
  He, Leahy, McDonell, Phang, Pieler, Prashanth, Purohit, Reynolds, Tow, Wang,
  and Weinbach}]{black-etal-2022-gpt}
Sidney Black, Stella Biderman, Eric Hallahan, Quentin Anthony, Leo Gao,
  Laurence Golding, Horace He, Connor Leahy, Kyle McDonell, Jason Phang,
  Michael Pieler, Usvsn~Sai Prashanth, Shivanshu Purohit, Laria Reynolds,
  Jonathan Tow, Ben Wang, and Samuel Weinbach. 2022.
\newblock \href {https://doi.org/10.18653/v1/2022.bigscience-1.9}
  {{GPT}-{N}eo{X}-20{B}: An open-source autoregressive language model}.
\newblock In \emph{Proceedings of BigScience Episode {\#}5 -- Workshop on
  Challenges {\&} Perspectives in Creating Large Language Models}, pages
  95--136, virtual+Dublin. Association for Computational Linguistics.

\bibitem[{Bojar et~al.(2016)Bojar, Du{\v{s}}ek, Kocmi, Libovick{\`y},
  Nov{\'a}k, Popel, Sudarikov, and Vari{\v{s}}}]{bojar2016czeng}
Ond{\v{r}}ej Bojar, Ond{\v{r}}ej Du{\v{s}}ek, Tom Kocmi, Jind{\v{r}}ich
  Libovick{\`y}, Michal Nov{\'a}k, Martin Popel, Roman Sudarikov, and
  Du{\v{s}}an Vari{\v{s}}. 2016.
\newblock Czeng 1.6: enlarged czech-english parallel corpus with processing
  tools dockered.
\newblock In \emph{Text, Speech, and Dialogue: 19th International Conference,
  TSD 2016, Brno, Czech Republic, September 12-16, 2016, Proceedings 19}, pages
  231--238. Springer.

\bibitem[{Brown et~al.(2020)Brown, Mann, Ryder, Subbiah, Kaplan, Dhariwal,
  Neelakantan, Shyam, Sastry, Askell, Agarwal, Herbert-Voss, Krueger, Henighan,
  Child, Ramesh, Ziegler, Wu, Winter, Hesse, Chen, Sigler, Litwin, Gray, Chess,
  Clark, Berner, McCandlish, Radford, Sutskever, and
  Amodei}]{brown-etal-2020-gpt3}
Tom Brown, Benjamin Mann, Nick Ryder, Melanie Subbiah, Jared~D Kaplan, Prafulla
  Dhariwal, Arvind Neelakantan, Pranav Shyam, Girish Sastry, Amanda Askell,
  Sandhini Agarwal, Ariel Herbert-Voss, Gretchen Krueger, Tom Henighan, Rewon
  Child, Aditya Ramesh, Daniel Ziegler, Jeffrey Wu, Clemens Winter, Chris
  Hesse, Mark Chen, Eric Sigler, Mateusz Litwin, Scott Gray, Benjamin Chess,
  Jack Clark, Christopher Berner, Sam McCandlish, Alec Radford, Ilya Sutskever,
  and Dario Amodei. 2020.
\newblock \href
  {https://proceedings.neurips.cc/paper_files/paper/2020/file/1457c0d6bfcb4967418bfb8ac142f64a-Paper.pdf}
  {Language models are few-shot learners}.
\newblock In \emph{Advances in Neural Information Processing Systems},
  volume~33, pages 1877--1901. Curran Associates, Inc.

\bibitem[{Capel(2012)}]{capel2012}
Annette Capel. 2012.
\newblock \href {https://doi.org/10.1017/S2041536212000013} {Completing the
  english vocabulary profile: C1 and c2 vocabulary}.
\newblock \emph{English Profile Journal}, 3:e1.

\bibitem[{Chen et~al.(2015)Chen, Huang, Chang, and
  Liou}]{nlplab-paraphrase-2015}
M.-H. Chen, S.-T. Huang, J.S. Chang, and H.-C. Liou. 2015.
\newblock \href {https://doi.org/10.1080/09588221.2013.783873} {Developing a
  corpus-based paraphrase tool to improve efl learners' writing skills}.
\newblock \emph{Computer Assisted Language Learning}, 28(1):22--40.

\bibitem[{Chung et~al.(2022)Chung, Hou, Longpre, Zoph, Tay, Fedus, Li, Wang,
  Dehghani, Brahma et~al.}]{chung2022scaling}
Hyung~Won Chung, Le~Hou, Shayne Longpre, Barret Zoph, Yi~Tay, William Fedus,
  Eric Li, Xuezhi Wang, Mostafa Dehghani, Siddhartha Brahma, et~al. 2022.
\newblock Scaling instruction-finetuned language models.
\newblock \emph{arXiv preprint arXiv:2210.11416}.

\bibitem[{Clive et~al.(2021)Clive, Cao, and Rei}]{clive2021control}
Jordan Clive, Kris Cao, and Marek Rei. 2021.
\newblock \href {https://doi.org/10.48550/ARXIV.2110.08329v2} {Control prefixes
  for text generation}.
\newblock \emph{arXiv preprint arXiv:2110.08329v2}.

\bibitem[{{Council of Europe}(2001)}]{council2001common}
{Council of Europe}. 2001.
\newblock \emph{Common European framework of reference for languages: Learning,
  teaching, assessment}.
\newblock Cambridge University Press.

\bibitem[{Devlin et~al.(2019)Devlin, Chang, Lee, and
  Toutanova}]{devlin-etal-2019-bert}
Jacob Devlin, Ming-Wei Chang, Kenton Lee, and Kristina Toutanova. 2019.
\newblock \href {https://doi.org/10.18653/v1/N19-1423} {{BERT}: Pre-training of
  deep bidirectional transformers for language understanding}.
\newblock In \emph{Proceedings of the 2019 Conference of the North {A}merican
  Chapter of the Association for Computational Linguistics: Human Language
  Technologies, Volume 1 (Long and Short Papers)}, pages 4171--4186,
  Minneapolis, Minnesota. Association for Computational Linguistics.

\bibitem[{Dmitrieva and Tiedemann(2020)}]{dmitrieva2020}
Anna Dmitrieva and J{\"o}rg Tiedemann. 2020.
\newblock \href {https://doi.org/10.1007/978-3-030-71214-3_7} {A multi-task
  learning approach to text simplification}.
\newblock In \emph{International Conference on Analysis of Images, Social
  Networks and Texts}, pages 78--89. Springer.

\bibitem[{Dong et~al.(2019)Dong, Li, Rezagholizadeh, and
  Cheung}]{dong-etal-2019-editnts}
Yue Dong, Zichao Li, Mehdi Rezagholizadeh, and Jackie Chi~Kit Cheung. 2019.
\newblock \href {https://doi.org/10.18653/v1/P19-1331} {{E}dit{NTS}: An neural
  programmer-interpreter model for sentence simplification through explicit
  editing}.
\newblock In \emph{Proceedings of the 57th Annual Meeting of the Association
  for Computational Linguistics}, pages 3393--3402, Florence, Italy.
  Association for Computational Linguistics.

\bibitem[{Feng et~al.(2023)Feng, Qiang, Li, Yuan, and Zhu}]{feng2023sentence}
Yutao Feng, Jipeng Qiang, Yun Li, Yunhao Yuan, and Yi~Zhu. 2023.
\newblock Sentence simplification via large language models.
\newblock \emph{arXiv preprint arXiv:2302.11957}.

\bibitem[{Fujinuma and Hagiwara(2021)}]{fujinuma2021cefr}
Yoshinari Fujinuma and Masato Hagiwara. 2021.
\newblock \href {https://doi.org/10.18653/v1/2021.textgraphs-1.16}
  {Semi-supervised joint estimation of word and document readability}.
\newblock In \emph{Proceedings of the Fifteenth Workshop on Graph-Based Methods
  for Natural Language Processing (TextGraphs-15)}, pages 150--155, Mexico
  City, Mexico. Association for Computational Linguistics.

\bibitem[{Gaillat et~al.(2022)Gaillat, Simpkin, Ballier, Stearns, Sousa,
  Bouyé, and Zarrouk}]{gaillat2022}
Thomas Gaillat, Andrew Simpkin, Nicolas Ballier, Bernardo Stearns, Annanda
  Sousa, Manon Bouyé, and Manel Zarrouk. 2022.
\newblock \href {https://doi.org/10.1017/S095834402100029X} {Predicting {CEFR}
  levels in learners of english: The use of microsystem criterial features in a
  machine learning approach}.
\newblock \emph{ReCALL}, 34(2):130–146.

\bibitem[{Gardner et~al.(2018)Gardner, Grus, Neumann, Tafjord, Dasigi, Liu,
  Peters, Schmitz, and Zettlemoyer}]{gardner-2018-allennlp}
Matt Gardner, Joel Grus, Mark Neumann, Oyvind Tafjord, Pradeep Dasigi,
  Nelson~F. Liu, Matthew Peters, Michael Schmitz, and Luke Zettlemoyer. 2018.
\newblock \href {https://doi.org/10.18653/v1/W18-2501} {{A}llen{NLP}: A deep
  semantic natural language processing platform}.
\newblock In \emph{Proceedings of Workshop for {NLP} Open Source Software
  ({NLP}-{OSS})}, pages 1--6, Melbourne, Australia. Association for
  Computational Linguistics.

\bibitem[{Guo et~al.(2018)Guo, Pasunuru, and Bansal}]{guo2018}
Han Guo, Ramakanth Pasunuru, and Mohit Bansal. 2018.
\newblock \href {https://aclanthology.org/C18-1039} {Dynamic multi-level
  multi-task learning for sentence simplification}.
\newblock In \emph{Proceedings of the 27th International Conference on
  Computational Linguistics}, pages 462--476, Santa Fe, New Mexico, USA.
  Association for Computational Linguistics.

\bibitem[{Holtzman et~al.(2019)Holtzman, Buys, Du, Forbes, and
  Choi}]{holtzman2019curious}
Ari Holtzman, Jan Buys, Li~Du, Maxwell Forbes, and Yejin Choi. 2019.
\newblock The curious case of neural text degeneration.
\newblock \emph{arXiv preprint arXiv:1904.09751}.

\bibitem[{Hu et~al.(2019)Hu, Rudinger, Post, and Van~Durme}]{hu2019parabank}
J~Edward Hu, Rachel Rudinger, Matt Post, and Benjamin Van~Durme. 2019.
\newblock Parabank: Monolingual bitext generation and sentential paraphrasing
  via lexically-constrained neural machine translation.
\newblock In \emph{Proceedings of the AAAI Conference on Artificial
  Intelligence}, volume~33, pages 6521--6528.

\bibitem[{Huang et~al.(2017)Huang, Chen, and Ku}]{example-suggester-2017}
Chieh-Yang Huang, Mei-Hua Chen, and Lun-Wei Ku. 2017.
\newblock \href {https://doi.org/10.1145/3041021.3054163} {Towards a better
  learning of near-synonyms: Automatically suggesting example sentences via
  fill in the blank}.
\newblock In \emph{Proceedings of the 26th International Conference on World
  Wide Web Companion}, WWW '17 Companion, page 293–302, Republic and Canton
  of Geneva, CHE. International World Wide Web Conferences Steering Committee.

\bibitem[{Iyer et~al.(2023)Iyer, Lin, Pasunuru, Mihaylov, Simig, Yu, Shuster,
  Wang, Liu, Koura, Li, O'Horo, Pereyra, Wang, Dewan, Celikyilmaz, Zettlemoyer,
  and Stoyanov}]{iyer2023optiml}
Srinivasan Iyer, Xi~Victoria Lin, Ramakanth Pasunuru, Todor Mihaylov, Daniel
  Simig, Ping Yu, Kurt Shuster, Tianlu Wang, Qing Liu, Punit~Singh Koura, Xian
  Li, Brian O'Horo, Gabriel Pereyra, Jeff Wang, Christopher Dewan, Asli
  Celikyilmaz, Luke Zettlemoyer, and Ves Stoyanov. 2023.
\newblock \href {http://arxiv.org/abs/2212.12017} {Opt-iml: Scaling language
  model instruction meta learning through the lens of generalization}.

\bibitem[{Jiang et~al.(2020)Jiang, Maddela, Lan, Zhong, and
  Xu}]{jiang-etal-2020-neural}
Chao Jiang, Mounica Maddela, Wuwei Lan, Yang Zhong, and Wei Xu. 2020.
\newblock \href {https://doi.org/10.18653/v1/2020.acl-main.709} {Neural {CRF}
  model for sentence alignment in text simplification}.
\newblock In \emph{Proceedings of the 58th Annual Meeting of the Association
  for Computational Linguistics}, pages 7943--7960, Online. Association for
  Computational Linguistics.

\bibitem[{Kerz et~al.(2021)Kerz, Wiechmann, Qiao, Tseng, and
  Str{\"o}bel}]{kerz2021}
Elma Kerz, Daniel Wiechmann, Yu~Qiao, Emma Tseng, and Marcus Str{\"o}bel. 2021.
\newblock \href {https://aclanthology.org/2021.bea-1.21} {Automated
  classification of written proficiency levels on the {CEFR}-scale through
  complexity contours and {RNN}s}.
\newblock In \emph{Proceedings of the 16th Workshop on Innovative Use of NLP
  for Building Educational Applications}, pages 199--209, Online. Association
  for Computational Linguistics.

\bibitem[{Kew and Ebling(2022)}]{kew-ebling-2022-target}
Tannon Kew and Sarah Ebling. 2022.
\newblock \href {https://aclanthology.org/2022.tsar-1.4} {Target-level sentence
  simplification as controlled paraphrasing}.
\newblock In \emph{Proceedings of the Workshop on Text Simplification,
  Accessibility, and Readability (TSAR-2022)}, pages 28--42, Abu Dhabi, United
  Arab Emirates (Virtual). Association for Computational Linguistics.

\bibitem[{Khallaf and Sharoff(2021)}]{khallaf2021}
Nouran Khallaf and Serge Sharoff. 2021.
\newblock \href {https://aclanthology.org/2021.wanlp-1.11} {Automatic
  difficulty classification of {A}rabic sentences}.
\newblock In \emph{Proceedings of the Sixth Arabic Natural Language Processing
  Workshop}, pages 105--114, Kyiv, Ukraine (Virtual). Association for
  Computational Linguistics.

\bibitem[{Kincaid et~al.(1975)Kincaid, Fishburne~Jr, Rogers, and
  Chissom}]{kincaid1975derivation}
J~Peter Kincaid, Robert~P Fishburne~Jr, Richard~L Rogers, and Brad~S Chissom.
  1975.
\newblock \href {https://apps.dtic.mil/sti/citations/ADA006655} {Derivation of
  new readability formulas (automated readability index, fog count and flesch
  reading ease formula) for navy enlisted personnel}.
\newblock Technical report, Naval Technical Training Command Millington TN
  Research Branch.

\bibitem[{Kingma and Ba(2015)}]{adam-kingma}
Diederik~P. Kingma and Jimmy Ba. 2015.
\newblock \href {http://arxiv.org/abs/1412.6980} {Adam: {A} method for
  stochastic optimization}.
\newblock In \emph{3rd International Conference on Learning Representations,
  {ICLR} 2015, San Diego, CA, USA, May 7-9, 2015, Conference Track
  Proceedings}.

\bibitem[{Krippendorff(2011)}]{krippendorff2011computing}
Klaus Krippendorff. 2011.
\newblock Computing krippendorff's alpha-reliability.

\bibitem[{Lee et~al.(2021)Lee, Jang, and Lee}]{lee-etal-2021-pushing}
Bruce~W. Lee, Yoo~Sung Jang, and Jason Lee. 2021.
\newblock \href {https://doi.org/10.18653/v1/2021.emnlp-main.834} {Pushing on
  text readability assessment: A transformer meets handcrafted linguistic
  features}.
\newblock In \emph{Proceedings of the 2021 Conference on Empirical Methods in
  Natural Language Processing}, pages 10669--10686, Online and Punta Cana,
  Dominican Republic. Association for Computational Linguistics.

\bibitem[{Lee and Vajjala(2022)}]{lee-vajjala-2022-neural}
Justin Lee and Sowmya Vajjala. 2022.
\newblock \href {https://doi.org/10.18653/v1/2022.findings-acl.300} {A neural
  pairwise ranking model for readability assessment}.
\newblock In \emph{Findings of the Association for Computational Linguistics:
  ACL 2022}, pages 3802--3813, Dublin, Ireland. Association for Computational
  Linguistics.

\bibitem[{Lewis et~al.(2020)Lewis, Liu, Goyal, Ghazvininejad, Mohamed, Levy,
  Stoyanov, and Zettlemoyer}]{lewis-etal-2020-bart}
Mike Lewis, Yinhan Liu, Naman Goyal, Marjan Ghazvininejad, Abdelrahman Mohamed,
  Omer Levy, Veselin Stoyanov, and Luke Zettlemoyer. 2020.
\newblock \href {https://doi.org/10.18653/v1/2020.acl-main.703} {{BART}:
  Denoising sequence-to-sequence pre-training for natural language generation,
  translation, and comprehension}.
\newblock In \emph{Proceedings of the 58th Annual Meeting of the Association
  for Computational Linguistics}, pages 7871--7880, Online. Association for
  Computational Linguistics.

\bibitem[{Liu et~al.(2019)Liu, He, Chen, and Gao}]{liu2019multi}
Xiaodong Liu, Pengcheng He, Weizhu Chen, and Jianfeng Gao. 2019.
\newblock \href {https://doi.org/10.18653/v1/P19-1441} {Multi-task deep neural
  networks for natural language understanding}.
\newblock In \emph{Proceedings of the 57th Annual Meeting of the Association
  for Computational Linguistics}, pages 4487--4496, Florence, Italy.
  Association for Computational Linguistics.

\bibitem[{Lu et~al.(2021)Lu, Qiang, Li, Yuan, and
  Zhu}]{lu-etal-2021-unsupervised-method}
Xinyu Lu, Jipeng Qiang, Yun Li, Yunhao Yuan, and Yi~Zhu. 2021.
\newblock \href {https://doi.org/10.18653/v1/2021.findings-emnlp.22} {An
  unsupervised method for building sentence simplification corpora in multiple
  languages}.
\newblock In \emph{Findings of the Association for Computational Linguistics:
  EMNLP 2021}, pages 227--237, Punta Cana, Dominican Republic. Association for
  Computational Linguistics.

\bibitem[{Maddela et~al.(2021)Maddela, Alva-Manchego, and
  Xu}]{maddela-etal-2021-controllable}
Mounica Maddela, Fernando Alva-Manchego, and Wei Xu. 2021.
\newblock \href {https://doi.org/10.18653/v1/2021.naacl-main.277} {Controllable
  text simplification with explicit paraphrasing}.
\newblock In \emph{Proceedings of the 2021 Conference of the North American
  Chapter of the Association for Computational Linguistics: Human Language
  Technologies}, pages 3536--3553, Online. Association for Computational
  Linguistics.

\bibitem[{Mallinson et~al.(2020)Mallinson, Sennrich, and
  Lapata}]{mallinson2020}
Jonathan Mallinson, Rico Sennrich, and Mirella Lapata. 2020.
\newblock \href {https://doi.org/10.18653/v1/2020.emnlp-main.415} {Zero-shot
  crosslingual sentence simplification}.
\newblock In \emph{Proceedings of the 2020 Conference on Empirical Methods in
  Natural Language Processing (EMNLP)}, pages 5109--5126, Online. Association
  for Computational Linguistics.

\bibitem[{Martin et~al.(2020{\natexlab{a}})Martin, Fan, de~la Clergerie,
  Bordes, and Sagot}]{martin2020access}
Louis Martin, Angela Fan, {\'E}ric de~la Clergerie, Antoine Bordes, and
  Beno{\^\i}t Sagot. 2020{\natexlab{a}}.
\newblock \href
  {http://114.215.220.151:8000/20200504/Multilingual\%20Unsupervised\%20Sentence\%20Simplification.pdf}
  {Multilingual unsupervised sentence simplification}.
\newblock \emph{arXiv preprint arXiv:2005.00352v1}.

\bibitem[{Martin et~al.(2020{\natexlab{b}})Martin, Fan, de~la Clergerie,
  Bordes, and Sagot}]{martin2020muss}
Louis Martin, Angela Fan, {\'E}ric de~la Clergerie, Antoine Bordes, and
  Beno{\^\i}t Sagot. 2020{\natexlab{b}}.
\newblock \href {https://doi.org/10.48550/arXiv.2005.00352} {{MUSS}:
  multilingual unsupervised sentence simplification by mining paraphrases}.
\newblock \emph{arXiv preprint arXiv:2005.00352v2}.

\bibitem[{Mikolov et~al.(2013)Mikolov, Sutskever, Chen, Corrado, and
  Dean}]{mikolov-word2vec-gnews}
Tomas Mikolov, Ilya Sutskever, Kai Chen, Greg~S Corrado, and Jeff Dean. 2013.
\newblock \href
  {https://proceedings.neurips.cc/paper_files/paper/2013/file/9aa42b31882ec039965f3c4923ce901b-Paper.pdf}
  {Distributed representations of words and phrases and their
  compositionality}.
\newblock In \emph{Advances in Neural Information Processing Systems},
  volume~26. Curran Associates, Inc.

\bibitem[{Narayan and Gardent(2014)}]{narayan2014hybrid}
Shashi Narayan and Claire Gardent. 2014.
\newblock Hybrid simplification using deep semantics and machine translation.
\newblock In \emph{The 52nd annual meeting of the association for computational
  linguistics}, pages 435--445.

\bibitem[{Naskar et~al.(2019)Naskar, Saha, and
  Mukherjee}]{naskar-etal-2019-text}
Subhajit Naskar, Soumya Saha, and Sreeparna Mukherjee. 2019.
\newblock \href {https://aclanthology.org/2019.ccnlg-1.4} {Text embellishment
  using attention based encoder-decoder model}.
\newblock In \emph{Proceedings of the 4th Workshop on Computational Creativity
  in Language Generation}, pages 28--38, Tokyo, Japan. Association for
  Computational Linguistics.

\bibitem[{Nishihara et~al.(2019)Nishihara, Kajiwara, and Arase}]{nishihara2019}
Daiki Nishihara, Tomoyuki Kajiwara, and Yuki Arase. 2019.
\newblock \href {https://doi.org/10.18653/v1/P19-2036} {Controllable text
  simplification with lexical constraint loss}.
\newblock In \emph{Proceedings of the 57th Annual Meeting of the Association
  for Computational Linguistics: Student Research Workshop}, pages 260--266,
  Florence, Italy. Association for Computational Linguistics.

\bibitem[{Raffel et~al.(2020)Raffel, Shazeer, Roberts, Lee, Narang, Matena,
  Zhou, Li, and Liu}]{raffelT5}
Colin Raffel, Noam Shazeer, Adam Roberts, Katherine Lee, Sharan Narang, Michael
  Matena, Yanqi Zhou, Wei Li, and Peter~J. Liu. 2020.
\newblock \href {http://jmlr.org/papers/v21/20-074.html} {Exploring the limits
  of transfer learning with a unified text-to-text transformer}.
\newblock \emph{Journal of Machine Learning Research}, 21(140):1--67.

\bibitem[{Ratner et~al.(2018)Ratner, Hancock, Dunnmon, Goldman, and
  R{\'e}}]{ratner2018snorkel}
Alex Ratner, Braden Hancock, Jared Dunnmon, Roger Goldman, and Christopher
  R{\'e}. 2018.
\newblock \href {https://doi.org/10.1145/3209889.3209898} {Snorkel metal: Weak
  supervision for multi-task learning}.
\newblock In \emph{Proceedings of the Second Workshop on Data Management for
  End-To-End Machine Learning}, pages 1--4.

\bibitem[{Reimers and Gurevych(2019)}]{reimers2019sbert}
Nils Reimers and Iryna Gurevych. 2019.
\newblock \href {https://doi.org/10.18653/v1/D19-1410} {Sentence-{BERT}:
  Sentence embeddings using {S}iamese {BERT}-networks}.
\newblock In \emph{Proceedings of the 2019 Conference on Empirical Methods in
  Natural Language Processing and the 9th International Joint Conference on
  Natural Language Processing (EMNLP-IJCNLP)}, pages 3982--3992, Hong Kong,
  China. Association for Computational Linguistics.

\bibitem[{Ryan et~al.(2023)Ryan, Naous, and Xu}]{Ryan_2023}
Michael~J. Ryan, Tarek Naous, and Wei Xu. 2023.
\newblock \href {https://arxiv.org/pdf/2305.15678.pdf} {Revisiting non-english
  text simplification: A unified multilingual benchmark}.
\newblock In \emph{ACL 2023}, pages 1--28.

\bibitem[{Scarton and Specia(2018)}]{scarton2018}
Carolina Scarton and Lucia Specia. 2018.
\newblock \href {https://doi.org/10.18653/v1/P18-2113} {Learning
  simplifications for specific target audiences}.
\newblock In \emph{Proceedings of the 56th Annual Meeting of the Association
  for Computational Linguistics (Volume 2: Short Papers)}, pages 712--718,
  Melbourne, Australia. Association for Computational Linguistics.

\bibitem[{Scherrer(2020)}]{scherrer2020tapaco}
Yves Scherrer. 2020.
\newblock Tapaco: A corpus of sentential paraphrases for 73 languages.
\newblock In \emph{Proceedings of the 12th Language Resources and Evaluation
  Conference}. European Language Resources Association (ELRA).

\bibitem[{Schmalz and Brutti(2021)}]{schmalz2021}
Veronica~Juliana Schmalz and Alessio Brutti. 2021.
\newblock \href {http://ceur-ws.org/Vol-3033/paper14.pdf} {Automatic assessment
  of english {CEFR} levels using {BERT} embeddings}.
\newblock \emph{http://ceur-ws.org/Vol-3033/}, 3033.

\bibitem[{Scholkopf et~al.(1997)Scholkopf, Sung, Burges, Girosi, Niyogi,
  Poggio, and Vapnik}]{scholkopf1997comparing}
Bernhard Scholkopf, Kah-Kay Sung, Christopher~JC Burges, Federico Girosi,
  Partha Niyogi, Tomaso Poggio, and Vladimir Vapnik. 1997.
\newblock Comparing support vector machines with gaussian kernels to radial
  basis function classifiers.
\newblock \emph{IEEE transactions on Signal Processing}, 45(11):2758--2765.

\bibitem[{Scialom et~al.(2022)Scialom, Chakrabarty, and
  Muresan}]{scialom-etal-2022-fine}
Thomas Scialom, Tuhin Chakrabarty, and Smaranda Muresan. 2022.
\newblock \href {https://aclanthology.org/2022.emnlp-main.410} {Fine-tuned
  language models are continual learners}.
\newblock In \emph{Proceedings of the 2022 Conference on Empirical Methods in
  Natural Language Processing}, pages 6107--6122, Abu Dhabi, United Arab
  Emirates. Association for Computational Linguistics.

\bibitem[{Settles et~al.(2020)Settles, LaFlair, and Hagiwara}]{settles2020}
Burr Settles, Geoffrey~T. LaFlair, and Masato Hagiwara. 2020.
\newblock \href {https://doi.org/10.1162/tacl_a_00310} {Machine
  learning{--}driven language assessment}.
\newblock \emph{Transactions of the Association for Computational Linguistics},
  8:247--263.

\bibitem[{Shin et~al.(2017)Shin, Lee, Kim, and Kim}]{rehearsal-shin-etal-2017}
Hanul Shin, Jung~Kwon Lee, Jaehong Kim, and Jiwon Kim. 2017.
\newblock Continual learning with deep generative replay.
\newblock In \emph{Proceedings of the 31st International Conference on Neural
  Information Processing Systems}, NIPS'17, page 2994–3003, Red Hook, NY,
  USA. Curran Associates Inc.

\bibitem[{Siddharthan(2002)}]{siddharthan2002architecture}
Advaith Siddharthan. 2002.
\newblock An architecture for a text simplification system.
\newblock In \emph{Language Engineering Conference, 2002. Proceedings}, pages
  64--71. IEEE.

\bibitem[{Siddharthan(2006)}]{siddharthan2006syntactic}
Advaith Siddharthan. 2006.
\newblock Syntactic simplification and text cohesion.
\newblock \emph{Research on Language and Computation}, 4:77--109.

\bibitem[{Sun et~al.(2023)Sun, Xu, and Wan}]{sun2023teaching}
Renliang Sun, Wei Xu, and Xiaojun Wan. 2023.
\newblock Teaching the pre-trained model to generate simple texts for text
  simplification.
\newblock \emph{arXiv preprint arXiv:2305.12463}.

\bibitem[{Tani et~al.(2022)Tani, Yuasa, Takikawa, Tamura, Kajiwara, Ninomiya,
  and Kato}]{tani-etal-2022-benchmark}
Kazuki Tani, Ryoya Yuasa, Kazuki Takikawa, Akihiro Tamura, Tomoyuki Kajiwara,
  Takashi Ninomiya, and Tsuneo Kato. 2022.
\newblock \href {https://aclanthology.org/2022.lrec-1.726} {A benchmark dataset
  for multi-level complexity-controllable machine translation}.
\newblock In \emph{Proceedings of the Thirteenth Language Resources and
  Evaluation Conference}, pages 6744--6752, Marseille, France. European
  Language Resources Association.

\bibitem[{Tanprasert and Kauchak(2021)}]{tanprasert-kauchak-2021-flesch}
Teerapaun Tanprasert and David Kauchak. 2021.
\newblock \href {https://doi.org/10.18653/v1/2021.gem-1.1} {Flesch-kincaid is
  not a text simplification evaluation metric}.
\newblock In \emph{Proceedings of the 1st Workshop on Natural Language
  Generation, Evaluation, and Metrics (GEM 2021)}, pages 1--14, Online.
  Association for Computational Linguistics.

\bibitem[{Tay et~al.(2023)Tay, Dehghani, Tran, Garcia, Wei, Wang, Chung, Bahri,
  Schuster, Zheng, Zhou, Houlsby, and Metzler}]{tay2023ul}
Yi~Tay, Mostafa Dehghani, Vinh~Q. Tran, Xavier Garcia, Jason Wei, Xuezhi Wang,
  Hyung~Won Chung, Dara Bahri, Tal Schuster, Steven Zheng, Denny Zhou, Neil
  Houlsby, and Donald Metzler. 2023.
\newblock \href {https://openreview.net/forum?id=6ruVLB727MC} {{UL}2: Unifying
  language learning paradigms}.
\newblock In \emph{The Eleventh International Conference on Learning
  Representations}.

\bibitem[{Thompson and Post(2020)}]{thompson-post-2020}
Brian Thompson and Matt Post. 2020.
\newblock \href {https://aclanthology.org/2020.wmt-1.67} {Paraphrase generation
  as zero-shot multilingual translation: Disentangling semantic similarity from
  lexical and syntactic diversity}.
\newblock In \emph{Proceedings of the Fifth Conference on Machine Translation},
  pages 561--570, Online. Association for Computational Linguistics.

\bibitem[{Tsai et~al.(2020)Tsai, Chen, Yang, and
  Chang}]{tsai-etal-2020-lingglewrite}
Chung-Ting Tsai, Jhih-Jie Chen, Ching-Yu Yang, and Jason~S. Chang. 2020.
\newblock \href {https://doi.org/10.18653/v1/2020.acl-demos.17}
  {{L}inggle{W}rite: a coaching system for essay writing}.
\newblock In \emph{Proceedings of the 58th Annual Meeting of the Association
  for Computational Linguistics: System Demonstrations}, pages 127--133,
  Online. Association for Computational Linguistics.

\bibitem[{Uchida et~al.(2018)Uchida, Takada, and Arase}]{uchida2018cefr}
Satoru Uchida, Shohei Takada, and Yuki Arase. 2018.
\newblock \href {https://aclanthology.org/L18-1514} {{CEFR}-based lexical
  simplification dataset}.
\newblock In \emph{Proceedings of the Eleventh International Conference on
  Language Resources and Evaluation ({LREC} 2018)}, Miyazaki, Japan. European
  Language Resources Association (ELRA).

\bibitem[{Vajjala and Lu{\v{c}}i{\'c}(2018)}]{2018-onestopenglish}
Sowmya Vajjala and Ivana Lu{\v{c}}i{\'c}. 2018.
\newblock \href {https://doi.org/10.18653/v1/W18-0535} {{O}ne{S}top{E}nglish
  corpus: A new corpus for automatic readability assessment and text
  simplification}.
\newblock In \emph{Proceedings of the Thirteenth Workshop on Innovative Use of
  {NLP} for Building Educational Applications}, pages 297--304, New Orleans,
  Louisiana. Association for Computational Linguistics.

\bibitem[{V{\'a}squez-Rodr{\'\i}guez et~al.(2022)V{\'a}squez-Rodr{\'\i}guez,
  Cuenca-Jim{\'e}nez, Morales-Esquivel, and
  Alva-Manchego}]{vasquez-rodriguez-etal-2022-benchmark}
Laura V{\'a}squez-Rodr{\'\i}guez, Pedro-Manuel Cuenca-Jim{\'e}nez, Sergio
  Morales-Esquivel, and Fernando Alva-Manchego. 2022.
\newblock \href {https://aclanthology.org/2022.tsar-1.18} {A benchmark for
  neural readability assessment of texts in {S}panish}.
\newblock In \emph{Proceedings of the Workshop on Text Simplification,
  Accessibility, and Readability (TSAR-2022)}, pages 188--198, Abu Dhabi,
  United Arab Emirates (Virtual). Association for Computational Linguistics.

\bibitem[{Volodina et~al.(2013)Volodina, Pijetlovic, Pil{\'a}n, and
  Kokkinakis}]{volodina2013}
Elena Volodina, Dijana Pijetlovic, Ildiko Pil{\'a}n, and Sofie~Johansson
  Kokkinakis. 2013.
\newblock \href
  {https://www.researchgate.net/publication/259173937_Towards_a_gold_standard_for_Swedish_CEFR-based_ICALL}
  {Towards a gold standard for swedish {CEFR}-based {ICALL}}.
\newblock In \emph{Proceedings of the Second Workshop on NLP for
  Computer-Assisted Language Learning. NEALT Proceedings Series}, volume~17.

\bibitem[{Welleck et~al.(2019)Welleck, Kulikov, Roller, Dinan, Cho, and
  Weston}]{welleck2019neural}
Sean Welleck, Ilia Kulikov, Stephen Roller, Emily Dinan, Kyunghyun Cho, and
  Jason Weston. 2019.
\newblock Neural text generation with unlikelihood training.
\newblock \emph{arXiv preprint arXiv:1908.04319}.

\bibitem[{Wieting and Gimpel(2018)}]{wieting-gimpel-2018-paranmt}
John Wieting and Kevin Gimpel. 2018.
\newblock \href {https://doi.org/10.18653/v1/P18-1042} {{P}ara{NMT}-50{M}:
  Pushing the limits of paraphrastic sentence embeddings with millions of
  machine translations}.
\newblock In \emph{Proceedings of the 56th Annual Meeting of the Association
  for Computational Linguistics (Volume 1: Long Papers)}, pages 451--462,
  Melbourne, Australia. Association for Computational Linguistics.

\bibitem[{Wolf et~al.(2020)Wolf, Debut, Sanh, Chaumond, Delangue, Moi, Cistac,
  Rault, Louf, Funtowicz, Davison, Shleifer, von Platen, Ma, Jernite, Plu, Xu,
  Scao, Gugger, Drame, Lhoest, and Rush}]{wolf-2020-transformers}
Thomas Wolf, Lysandre Debut, Victor Sanh, Julien Chaumond, Clement Delangue,
  Anthony Moi, Pierric Cistac, Tim Rault, Rémi Louf, Morgan Funtowicz, Joe
  Davison, Sam Shleifer, Patrick von Platen, Clara Ma, Yacine Jernite, Julien
  Plu, Canwen Xu, Teven~Le Scao, Sylvain Gugger, Mariama Drame, Quentin Lhoest,
  and Alexander~M. Rush. 2020.
\newblock \href {https://www.aclweb.org/anthology/2020.emnlp-demos.6}
  {Transformers: State-of-the-art natural language processing}.
\newblock In \emph{Proceedings of the 2020 Conference on Empirical Methods in
  Natural Language Processing: System Demonstrations}, pages 38--45, Online.
  Association for Computational Linguistics.

\bibitem[{Woodsend and Lapata(2011)}]{woodsend2011learning}
Kristian Woodsend and Mirella Lapata. 2011.
\newblock Learning to simplify sentences with quasi-synchronous grammar and
  integer programming.
\newblock In \emph{Proceedings of the 2011 Conference on Empirical Methods in
  Natural Language Processing}, pages 409--420.

\bibitem[{Workshop et~al.(2023)Workshop, :, Scao, Fan, Akiki, Pavlick, Ilić,
  Hesslow, Castagné, Luccioni, Yvon, Gallé, Tow, Rush, Biderman, Webson,
  Ammanamanchi, Wang, Sagot, Muennighoff, del Moral, Ruwase, Bawden, Bekman,
  McMillan-Major, Beltagy, Nguyen, Saulnier, Tan, Suarez, Sanh, Laurençon,
  Jernite, Launay, Mitchell, Raffel, Gokaslan, Simhi, Soroa, Aji, Alfassy,
  Rogers, Nitzav, Xu, Mou, Emezue, Klamm, Leong, van Strien, Adelani, Radev,
  Ponferrada, Levkovizh, Kim, Natan, Toni, Dupont, Kruszewski, Pistilli,
  Elsahar, Benyamina, Tran, Yu, Abdulmumin, Johnson, Gonzalez-Dios, de~la Rosa,
  Chim, Dodge, Zhu, Chang, Frohberg, Tobing, Bhattacharjee, Almubarak, Chen,
  Lo, Werra, Weber, Phan, allal, Tanguy, Dey, Muñoz, Masoud, Grandury,
  Šaško, Huang, Coavoux, Singh, Jiang, Vu, Jauhar, Ghaleb, Subramani,
  Kassner, Khamis, Nguyen, Espejel, de~Gibert, Villegas, Henderson, Colombo,
  Amuok, Lhoest, Harliman, Bommasani, López, Ribeiro, Osei, Pyysalo, Nagel,
  Bose, Muhammad, Sharma, Longpre, Nikpoor, Silberberg, Pai, Zink, Torrent,
  Schick, Thrush, Danchev, Nikoulina, Laippala, Lepercq, Prabhu, Alyafeai,
  Talat, Raja, Heinzerling, Si, Taşar, Salesky, Mielke, Lee, Sharma, Santilli,
  Chaffin, Stiegler, Datta, Szczechla, Chhablani, Wang, Pandey, Strobelt,
  Fries, Rozen, Gao, Sutawika, Bari, Al-shaibani, Manica, Nayak, Teehan,
  Albanie, Shen, Ben-David, Bach, Kim, Bers, Fevry, Neeraj, Thakker, Raunak,
  Tang, Yong, Sun, Brody, Uri, Tojarieh, Roberts, Chung, Tae, Phang, Press, Li,
  Narayanan, Bourfoune, Casper, Rasley, Ryabinin, Mishra, Zhang, Shoeybi,
  Peyrounette, Patry, Tazi, Sanseviero, von Platen, Cornette, Lavallée,
  Lacroix, Rajbhandari, Gandhi, Smith, Requena, Patil, Dettmers, Baruwa, Singh,
  Cheveleva, Ligozat, Subramonian, Névéol, Lovering, Garrette, Tunuguntla,
  Reiter, Taktasheva, Voloshina, Bogdanov, Winata, Schoelkopf, Kalo, Novikova,
  Forde, Clive, Kasai, Kawamura, Hazan, Carpuat, Clinciu, Kim, Cheng, Serikov,
  Antverg, van~der Wal, Zhang, Zhang, Gehrmann, Mirkin, Pais, Shavrina,
  Scialom, Yun, Limisiewicz, Rieser, Protasov, Mikhailov, Pruksachatkun,
  Belinkov, Bamberger, Kasner, Rueda, Pestana, Feizpour, Khan, Faranak, Santos,
  Hevia, Unldreaj, Aghagol, Abdollahi, Tammour, HajiHosseini, Behroozi,
  Ajibade, Saxena, Ferrandis, McDuff, Contractor, Lansky, David, Kiela, Nguyen,
  Tan, Baylor, Ozoani, Mirza, Ononiwu, Rezanejad, Jones, Bhattacharya,
  Solaiman, Sedenko, Nejadgholi, Passmore, Seltzer, Sanz, Dutra, Samagaio,
  Elbadri, Mieskes, Gerchick, Akinlolu, McKenna, Qiu, Ghauri, Burynok, Abrar,
  Rajani, Elkott, Fahmy, Samuel, An, Kromann, Hao, Alizadeh, Shubber, Wang,
  Roy, Viguier, Le, Oyebade, Le, Yang, Nguyen, Kashyap, Palasciano, Callahan,
  Shukla, Miranda-Escalada, Singh, Beilharz, Wang, Brito, Zhou, Jain, Xu,
  Fourrier, Periñán, Molano, Yu, Manjavacas, Barth, Fuhrimann, Altay, Bayrak,
  Burns, Vrabec, Bello, Dash, Kang, Giorgi, Golde, Posada, Sivaraman,
  Bulchandani, Liu, Shinzato, de~Bykhovetz, Takeuchi, Pàmies, Castillo,
  Nezhurina, Sänger, Samwald, Cullan, Weinberg, Wolf, Mihaljcic, Liu,
  Freidank, Kang, Seelam, Dahlberg, Broad, Muellner, Fung, Haller,
  Chandrasekhar, Eisenberg, Martin, Canalli, Su, Su, Cahyawijaya, Garda,
  Deshmukh, Mishra, Kiblawi, Ott, Sang-aroonsiri, Kumar, Schweter, Bharati,
  Laud, Gigant, Kainuma, Kusa, Labrak, Bajaj, Venkatraman, Xu, Xu, Xu, Tan,
  Xie, Ye, Bras, Belkada, and Wolf}]{workshop2023bloom}
BigScience Workshop, :, Teven~Le Scao, Angela Fan, Christopher Akiki, Ellie
  Pavlick, Suzana Ilić, Daniel Hesslow, Roman Castagné, Alexandra~Sasha
  Luccioni, François Yvon, Matthias Gallé, Jonathan Tow, Alexander~M. Rush,
  Stella Biderman, Albert Webson, Pawan~Sasanka Ammanamanchi, Thomas Wang,
  Benoît Sagot, Niklas Muennighoff, Albert~Villanova del Moral, Olatunji
  Ruwase, Rachel Bawden, Stas Bekman, Angelina McMillan-Major, Iz~Beltagy, Huu
  Nguyen, Lucile Saulnier, Samson Tan, Pedro~Ortiz Suarez, Victor Sanh, Hugo
  Laurençon, Yacine Jernite, Julien Launay, Margaret Mitchell, Colin Raffel,
  Aaron Gokaslan, Adi Simhi, Aitor Soroa, Alham~Fikri Aji, Amit Alfassy, Anna
  Rogers, Ariel~Kreisberg Nitzav, Canwen Xu, Chenghao Mou, Chris Emezue,
  Christopher Klamm, Colin Leong, Daniel van Strien, David~Ifeoluwa Adelani,
  Dragomir Radev, Eduardo~González Ponferrada, Efrat Levkovizh, Ethan Kim,
  Eyal~Bar Natan, Francesco~De Toni, Gérard Dupont, Germán Kruszewski, Giada
  Pistilli, Hady Elsahar, Hamza Benyamina, Hieu Tran, Ian Yu, Idris Abdulmumin,
  Isaac Johnson, Itziar Gonzalez-Dios, Javier de~la Rosa, Jenny Chim, Jesse
  Dodge, Jian Zhu, Jonathan Chang, Jörg Frohberg, Joseph Tobing, Joydeep
  Bhattacharjee, Khalid Almubarak, Kimbo Chen, Kyle Lo, Leandro~Von Werra, Leon
  Weber, Long Phan, Loubna~Ben allal, Ludovic Tanguy, Manan Dey, Manuel~Romero
  Muñoz, Maraim Masoud, María Grandury, Mario Šaško, Max Huang, Maximin
  Coavoux, Mayank Singh, Mike Tian-Jian Jiang, Minh~Chien Vu, Mohammad~A.
  Jauhar, Mustafa Ghaleb, Nishant Subramani, Nora Kassner, Nurulaqilla Khamis,
  Olivier Nguyen, Omar Espejel, Ona de~Gibert, Paulo Villegas, Peter Henderson,
  Pierre Colombo, Priscilla Amuok, Quentin Lhoest, Rheza Harliman, Rishi
  Bommasani, Roberto~Luis López, Rui Ribeiro, Salomey Osei, Sampo Pyysalo,
  Sebastian Nagel, Shamik Bose, Shamsuddeen~Hassan Muhammad, Shanya Sharma,
  Shayne Longpre, Somaieh Nikpoor, Stanislav Silberberg, Suhas Pai, Sydney
  Zink, Tiago~Timponi Torrent, Timo Schick, Tristan Thrush, Valentin Danchev,
  Vassilina Nikoulina, Veronika Laippala, Violette Lepercq, Vrinda Prabhu, Zaid
  Alyafeai, Zeerak Talat, Arun Raja, Benjamin Heinzerling, Chenglei Si,
  Davut~Emre Taşar, Elizabeth Salesky, Sabrina~J. Mielke, Wilson~Y. Lee,
  Abheesht Sharma, Andrea Santilli, Antoine Chaffin, Arnaud Stiegler, Debajyoti
  Datta, Eliza Szczechla, Gunjan Chhablani, Han Wang, Harshit Pandey, Hendrik
  Strobelt, Jason~Alan Fries, Jos Rozen, Leo Gao, Lintang Sutawika, M~Saiful
  Bari, Maged~S. Al-shaibani, Matteo Manica, Nihal Nayak, Ryan Teehan, Samuel
  Albanie, Sheng Shen, Srulik Ben-David, Stephen~H. Bach, Taewoon Kim, Tali
  Bers, Thibault Fevry, Trishala Neeraj, Urmish Thakker, Vikas Raunak, Xiangru
  Tang, Zheng-Xin Yong, Zhiqing Sun, Shaked Brody, Yallow Uri, Hadar Tojarieh,
  Adam Roberts, Hyung~Won Chung, Jaesung Tae, Jason Phang, Ofir Press, Conglong
  Li, Deepak Narayanan, Hatim Bourfoune, Jared Casper, Jeff Rasley, Max
  Ryabinin, Mayank Mishra, Minjia Zhang, Mohammad Shoeybi, Myriam Peyrounette,
  Nicolas Patry, Nouamane Tazi, Omar Sanseviero, Patrick von Platen, Pierre
  Cornette, Pierre~François Lavallée, Rémi Lacroix, Samyam Rajbhandari,
  Sanchit Gandhi, Shaden Smith, Stéphane Requena, Suraj Patil, Tim Dettmers,
  Ahmed Baruwa, Amanpreet Singh, Anastasia Cheveleva, Anne-Laure Ligozat, Arjun
  Subramonian, Aurélie Névéol, Charles Lovering, Dan Garrette, Deepak
  Tunuguntla, Ehud Reiter, Ekaterina Taktasheva, Ekaterina Voloshina, Eli
  Bogdanov, Genta~Indra Winata, Hailey Schoelkopf, Jan-Christoph Kalo,
  Jekaterina Novikova, Jessica~Zosa Forde, Jordan Clive, Jungo Kasai, Ken
  Kawamura, Liam Hazan, Marine Carpuat, Miruna Clinciu, Najoung Kim, Newton
  Cheng, Oleg Serikov, Omer Antverg, Oskar van~der Wal, Rui Zhang, Ruochen
  Zhang, Sebastian Gehrmann, Shachar Mirkin, Shani Pais, Tatiana Shavrina,
  Thomas Scialom, Tian Yun, Tomasz Limisiewicz, Verena Rieser, Vitaly Protasov,
  Vladislav Mikhailov, Yada Pruksachatkun, Yonatan Belinkov, Zachary Bamberger,
  Zdeněk Kasner, Alice Rueda, Amanda Pestana, Amir Feizpour, Ammar Khan, Amy
  Faranak, Ana Santos, Anthony Hevia, Antigona Unldreaj, Arash Aghagol, Arezoo
  Abdollahi, Aycha Tammour, Azadeh HajiHosseini, Bahareh Behroozi, Benjamin
  Ajibade, Bharat Saxena, Carlos~Muñoz Ferrandis, Daniel McDuff, Danish
  Contractor, David Lansky, Davis David, Douwe Kiela, Duong~A. Nguyen, Edward
  Tan, Emi Baylor, Ezinwanne Ozoani, Fatima Mirza, Frankline Ononiwu, Habib
  Rezanejad, Hessie Jones, Indrani Bhattacharya, Irene Solaiman, Irina Sedenko,
  Isar Nejadgholi, Jesse Passmore, Josh Seltzer, Julio~Bonis Sanz, Livia Dutra,
  Mairon Samagaio, Maraim Elbadri, Margot Mieskes, Marissa Gerchick, Martha
  Akinlolu, Michael McKenna, Mike Qiu, Muhammed Ghauri, Mykola Burynok, Nafis
  Abrar, Nazneen Rajani, Nour Elkott, Nour Fahmy, Olanrewaju Samuel, Ran An,
  Rasmus Kromann, Ryan Hao, Samira Alizadeh, Sarmad Shubber, Silas Wang, Sourav
  Roy, Sylvain Viguier, Thanh Le, Tobi Oyebade, Trieu Le, Yoyo Yang, Zach
  Nguyen, Abhinav~Ramesh Kashyap, Alfredo Palasciano, Alison Callahan, Anima
  Shukla, Antonio Miranda-Escalada, Ayush Singh, Benjamin Beilharz, Bo~Wang,
  Caio Brito, Chenxi Zhou, Chirag Jain, Chuxin Xu, Clémentine Fourrier,
  Daniel~León Periñán, Daniel Molano, Dian Yu, Enrique Manjavacas, Fabio
  Barth, Florian Fuhrimann, Gabriel Altay, Giyaseddin Bayrak, Gully Burns,
  Helena~U. Vrabec, Imane Bello, Ishani Dash, Jihyun Kang, John Giorgi, Jonas
  Golde, Jose~David Posada, Karthik~Rangasai Sivaraman, Lokesh Bulchandani,
  Lu~Liu, Luisa Shinzato, Madeleine~Hahn de~Bykhovetz, Maiko Takeuchi, Marc
  Pàmies, Maria~A Castillo, Marianna Nezhurina, Mario Sänger, Matthias
  Samwald, Michael Cullan, Michael Weinberg, Michiel~De Wolf, Mina Mihaljcic,
  Minna Liu, Moritz Freidank, Myungsun Kang, Natasha Seelam, Nathan Dahlberg,
  Nicholas~Michio Broad, Nikolaus Muellner, Pascale Fung, Patrick Haller, Ramya
  Chandrasekhar, Renata Eisenberg, Robert Martin, Rodrigo Canalli, Rosaline Su,
  Ruisi Su, Samuel Cahyawijaya, Samuele Garda, Shlok~S Deshmukh, Shubhanshu
  Mishra, Sid Kiblawi, Simon Ott, Sinee Sang-aroonsiri, Srishti Kumar, Stefan
  Schweter, Sushil Bharati, Tanmay Laud, Théo Gigant, Tomoya Kainuma, Wojciech
  Kusa, Yanis Labrak, Yash~Shailesh Bajaj, Yash Venkatraman, Yifan Xu, Yingxin
  Xu, Yu~Xu, Zhe Tan, Zhongli Xie, Zifan Ye, Mathilde Bras, Younes Belkada, and
  Thomas Wolf. 2023.
\newblock \href {http://arxiv.org/abs/2211.05100} {Bloom: A 176b-parameter
  open-access multilingual language model}.

\bibitem[{Wubben et~al.(2012)Wubben, Van Den~Bosch, and
  Krahmer}]{wubben2012sentence}
Sander Wubben, Antal Van Den~Bosch, and Emiel Krahmer. 2012.
\newblock Sentence simplification by monolingual machine translation.
\newblock In \emph{Proceedings of the 50th Annual Meeting of the Association
  for Computational Linguistics (Volume 1: Long Papers)}, pages 1015--1024.

\bibitem[{Xia et~al.(2016)Xia, Kochmar, and Briscoe}]{xia-etal-2016-text}
Menglin Xia, Ekaterina Kochmar, and Ted Briscoe. 2016.
\newblock \href {https://doi.org/10.18653/v1/W16-0502} {Text readability
  assessment for second language learners}.
\newblock In \emph{Proceedings of the 11th Workshop on Innovative Use of {NLP}
  for Building Educational Applications}, pages 12--22, San Diego, CA.
  Association for Computational Linguistics.

\bibitem[{Xu et~al.(2015)Xu, Callison-Burch, and Napoles}]{tacl-newsela}
Wei Xu, Chris Callison-Burch, and Courtney Napoles. 2015.
\newblock \href {https://doi.org/10.1162/tacl_a_00139} {{Problems in Current
  Text Simplification Research: New Data Can Help}}.
\newblock \emph{Transactions of the Association for Computational Linguistics},
  3:283--297.

\bibitem[{Xu et~al.(2016)Xu, Napoles, Pavlick, Chen, and
  Callison-Burch}]{xu2016sari}
Wei Xu, Courtney Napoles, Ellie Pavlick, Quanze Chen, and Chris Callison-Burch.
  2016.
\newblock \href {https://doi.org/10.1162/tacl_a_00107} {Optimizing statistical
  machine translation for text simplification}.
\newblock \emph{Transactions of the Association for Computational Linguistics},
  4:401--415.

\bibitem[{Zhang and Lapata(2017)}]{zhangLapata2017}
Xingxing Zhang and Mirella Lapata. 2017.
\newblock \href {https://doi.org/10.18653/v1/D17-1062} {Sentence simplification
  with deep reinforcement learning}.
\newblock In \emph{Proceedings of the 2017 Conference on Empirical Methods in
  Natural Language Processing}, pages 584--594, Copenhagen, Denmark.
  Association for Computational Linguistics.

\bibitem[{Zhao(2022)}]{zhao2022leveraging}
Xin Zhao. 2022.
\newblock Leveraging artificial intelligence (ai) technology for english
  writing: Introducing wordtune as a digital writing assistant for efl writers.
\newblock \emph{RELC Journal}, page 00336882221094089.

\bibitem[{Zhu et~al.(2010)Zhu, Bernhard, and
  Gurevych}]{zhu-etal-2010-monolingual}
Zhemin Zhu, Delphine Bernhard, and Iryna Gurevych. 2010.
\newblock \href {https://aclanthology.org/C10-1152} {A monolingual tree-based
  translation model for sentence simplification}.
\newblock In \emph{Proceedings of the 23rd International Conference on
  Computational Linguistics (Coling 2010)}, pages 1353--1361, Beijing, China.
  Coling 2010 Organizing Committee.

\end{thebibliography}
\bibliographystyle{acl_natbib}

\iftaclpubformat

\onecolumn
  
\fi

\end{document}